\documentclass[letterpaper]{article}
\pdfoutput=1 
\usepackage{amsmath}
\usepackage{graphicx}
\usepackage{subfigure}

\author{Dongwoo Kim, Yeonseung Chung, and Alice Oh
\\ KAIST, Korea} 
\title{Variable Selection for Latent Dirichlet Allocation}

\begin{document}
\maketitle

\begin{abstract}
\begin{quote}
In latent Dirichlet allocation (LDA), topics are multinomial distributions over the entire vocabulary. However, the vocabulary usually contains many words that are not relevant in forming the topics. We adopt a variable selection method widely used in statistical modeling as a dimension reduction tool and combine it with LDA. In this variable selection model for LDA (vsLDA), topics are multinomial distributions over a subset of the vocabulary, and by excluding words that are not informative for finding the latent topic structure of the corpus, vsLDA finds topics that are more robust and discriminative. We compare three models, vsLDA, LDA with symmetric priors, and LDA with asymmetric priors, on heldout likelihood, MCMC chain consistency, and document classification. The performance of vsLDA is better than symmetric LDA for likelihood and classification, better than asymmetric LDA for consistency and classification, and about the same in the other comparisons.
\end{quote}
\end{abstract}

\section{Introduction}

Latent Dirichlet allocation (LDA) \cite{Blei:2003p4796}, a widely used topic model, decomposes a corpus into a finite set of topics. Each topic is a multinomial distribution over the entire vocabulary, which is typically defined to be the set of all unique words with an optional step of removing stopwords and high frequency words.  Even with the preprocessing step, the vocabulary will almost certainly contain words that do not contribute to the underlying topical structure of the corpus, and those words may interfere with the model's ability to find topics with predictive and discriminative power. More importantly, one cannot be sure whether and how much the vocabulary influences the topics inferred, and there is not a systematic way to compare different vocabularies for a given corpus. We relax the constraint that the vocabulary must be fixed a priori and let the topic model consider any subset of the vocabulary for representing the topics. 

We propose a model-based variable selection \cite{citeulike:4523099,RefWorks:12} for LDA (vsLDA) that combines the process of identifying a relevant subset of the vocabulary with the process of finding the topics. 
Variable selection has not been studied in depth for LDA or any other topic model, but three models, HMM-LDA \cite{citeulike:4302386}, sparseTM
\cite{Wang:2010p3585}, and SWB \cite{chemudugunta2007modeling} achieve
a similar effect of representing the topics with a subset of the
vocabulary.
HMM-LDA \cite{citeulike:4302386} models the short- and long-range dependencies of the words and thus identifies whether words are generated from the syntactic (non-topic) or the semantic (topic) class. SparseTM \cite{Wang:2010p3585} aims to decouple sparsity and smoothness of the word-topic distribution and thereby excludes some words from each topic. SWB seperates \textit{word tokens} into the general and specific aspects, and it is probably the most similar work to ours in that it also globally excludes words from forming the topics. However, SWB excludes {\it word tokens}, whereas vsLDA excludes {\it word types}. By looking at the word types, we can replace the necessary but arbitrary step of deciding the vocabulary for forming the topics, which usually includes the removal process of useless words. Such process typically uses a list of stop words and corpus-dependent infrequent and highly frequent words, and in this work, we show the inadequacy of such preprocessing approach to variable selection. We can also view this problem of variable selection as a type of model selection along the vocabulary dimension. Model selection has been well studied for the topic dimension with nonparametric topic models \cite{Teh:2006p3792,Wang:2010p3585} but not for the vocabulary dimension.

This paper is organized as follows. We first describe our vsLDA model for selecting informative words. We derive an approximate algorithm for posterior inference on the latent variables of interest in vsLDA based on Markov Chain Monte Carlo and Monte Carlo integration. We demonstrate our approach on a synthetic dataset to verify the correctness of our model. Then we run our model on three real-world datasets and compare the performance with LDA with symmetric priors (symLDA) and LDA with asymmetric priors (asymLDA). We show that vsLDA finds topics with better predictive power than symLDA and more robustness than asymLDA. We also find that vsLDA reduces each document into more discriminating subdimensions and hence outperforms the other models for document classification.

\section{Variable Selection for LDA (vsLDA)}
LDA is typically used with a preprocessing step of removing stopwords and the words that occur frequently throughout the corpus. The rationale is that the words pervading the corpus do not contribute to but hinder the process of discovering a latent topic structure. This frequency-based preprocessing step excludes the words a priori independent of constructing the latent topics. However, we cannot be certain whether the excluded words are truly non-informative for topic construction. Also, the same uncertainty applies to the included words. Here, we propose a new LDA model where the word selection is conducted simultaneously while discovering the latent topics. The proposed approach combines a stochastic search variable selection \cite{citeulike:4523099} with LDA, providing an automatic word selection procedure for topic models. 

Suppose we have a vocabulary with size $V$ with or without any preprocessing. In a typical topic model, topics are defined on the entire vocabulary and assumed to be Dirichlet-distributed on $V-1$ simplex, i.e.,
\begin{align}
\phi_k \sim \mbox{Dir}(\beta\mathbf{1}),  \qquad{}k \in \{1,2,3, \ldots, K \},
\end{align}
where $\mathbf{1}$ is $V$-dimensional vector of 1s and $K$ is the number of topics. Our assumption is that the vocabulary is divided into two mutually exclusive word sets; one includes informative words for constructing topics, and the other contains non-informative words. Also, the topics are assumed to be defined only on the informative word set and distributed as 
\begin{align}
\phi_k \sim \mbox{Dir}(\beta\mathbf{s}), \qquad{}k \in \{1,2,3, \ldots, K \},
\end{align}
where $\mathbf{s} = (s_1, \ldots, s_V)$ and $s_j$ is an indicator variable defined as
\begin{eqnarray}
s_j = \left\{ \begin{array}{ll}
1, &\mbox{word $j$ is a informative word,}\\
0, &\mbox{word $j$ is a non-informative word.}
       \end{array} \right.
\end{eqnarray}
In other words, $\mathbf{s}$ specifies a smaller simplex with a dimension $\sum_{j=1}^V s_j - 1$ for the informative word set. Not knowing a priori whether a word is informative or non-informative, we assume $s_j \sim Bernoulli (\lambda)$ to incorporate uncertainty in informativity of words. 

Now, we describe the generative process for vsLDA which includes the steps for dividing the entire vocabulary into an informative word set and a non-informative word set (step 1) and determining the membership of a word token either as one of the topics or as the non-informative word set (step 4(b)).
\begin{enumerate}
\item{For each word $j \in \{1,2,\ldots,V\}$, draw word selection variable $s_{j} \sim \mbox{Bernoulli}(\lambda)$}
\item{For each topic $k \in \{1,2,\ldots,K\}$, draw topic distribution $\phi_k \sim \mbox{Dir}(\beta \boldsymbol{s})$ }
\item{For a non-informative words set, draw words distribution $\psi \sim \mbox{Dir}(\gamma \boldsymbol{s}^c)$}
\item{For each document $d \in \{1,2,\ldots,D\}$:}
\begin{enumerate}
\item{Draw topic proportion $\theta_d \sim \mbox{Dir}(\alpha)$}
\item{For $i$th word token, draw $b_{di} \sim \mbox{Bernoulli}(\tau)$:}
\begin{enumerate}
\item{If $b_{di} = 1$:}
\begin{enumerate}
\item{Draw topic assignment $z_{di} \sim \mbox{Mult}(\theta_d)$}
\item{Draw word token $w_{di} \sim \mbox{Mult}(\phi_{z_{di}})$}
\end{enumerate}
\item{else}
\begin{enumerate}
\item{Draw word token $w_{di} \sim \mbox{Mult}(\psi)$}
\end{enumerate}
\end{enumerate}
\end{enumerate}
\end{enumerate}


From the generative process for a corpus, the likelihood of the corpus is
\begin{align*}
&P(W, \Theta, \Phi, \psi, \boldsymbol{z}, \boldsymbol{b}, \boldsymbol{s} | \alpha,\beta,\gamma, \lambda, \tau)  &\notag\\
&= \prod_{d=1}^D p(\theta_d|\alpha)  \prod_{d=1}^D \prod_{\{i : b_{di} = 1\}} \{ p(w_{di}|\phi_{z_{di}}) p(z_{di}|\theta_d) \} &\notag\\
&\times \prod_{d=1}^D \prod_{\{ i : b_{di} = 0\}} p(w_{di}|\psi) \prod_{d=1}^D \prod_{i=1}^{N_d}p(b_{di}|\tau) &\notag\\
&\times \prod_{k=1}^K p(\phi_k|\beta, {\bf s}) p(\psi|\gamma, {\bf s}) \prod_{j=1}^V p(s_j|\lambda),&
\end{align*}
\normalsize
where $N_d$ is the number of word tokens in document $d$.

Placing Dirichlet-multinomial conjugate priors over $\Theta, \Phi, \psi$ naturally leads to marginalizing out these variables.
\begin{align}
\label{eqnm}
&p(W, \boldsymbol{z}, \boldsymbol{b}, \boldsymbol{s}| \alpha, \beta, \gamma, \lambda, \tau) = &\notag\\
&\int_{\Theta} \prod_{d=1}^D \prod_{ \{i : b_{di} = 1\}} p(z_{di}|\theta_d) p(\theta_d|\alpha) d\Theta  &\notag\\
& \times \int_{\Phi} \prod_{k=1}^K p(\phi_k|\beta, {\bf s}) \prod_{d=1}^D \prod_{\{i : b_{di} = 1\}} p(w_{di}|\phi_{z_{di}}) d\Phi &\notag\\
&\times \int_{\psi} p(\psi|\gamma, {\bf s}) \prod_{d=1}^D \prod_{\{i : b_{di} = 0\}} p(w_{di} | \psi) d\psi   &\notag\\
&\times \prod_{j=1}^V p(s_j|\lambda)  \prod_{d=1}^D \prod_{i=1}^{N_d} p(b_{di} | \tau) &\\
&= \prod_{d=1}^D \frac{ \Gamma(\sum_{k=1}^K \alpha_k) }{ \prod_{k=1}^K \Gamma(\alpha_k) } 
\frac{\prod_{k=1}^K \Gamma(n_{d\cdot}^k + \alpha_k)}{ \Gamma( \sum_{k=1}^K n_{d\cdot}^k  + \alpha_k )}  &\notag\\
&\times\prod_{k=1}^K \frac{\Gamma(\sum_{\{j : s_{j} = 1\}} \beta_j)} { \prod_{\{j : s_{j} = 1\}} \Gamma(\beta_j)}
\frac{\prod_{\{j : s_{j} = 1\}} \Gamma(n^k_{\cdot j} + \beta_j) }{\Gamma(\sum_{\{j : s_{j} = 1\}}n^k_{\cdot j} + \beta_j)} &\notag\\
&\times \frac{\Gamma(\sum_{\{j : s_{j} = 0\}} \gamma_j)} { \prod_{\{j : s_{j} = 0\}} \Gamma(\gamma_j)} 
\frac{\prod_{\{j : s_{j} = 0\}} \Gamma(m_{\cdot j} + \gamma_j) }{\Gamma(\sum_{\{j : s_{j} = 0\}}m_{\cdot j} +  \gamma_j)}  &\notag\\
& \times \lambda^{|\boldsymbol{s}|}  (1-\lambda)^{|V|-|\boldsymbol{s}|} \cdot
\tau^{n_{\cdot\cdot}^{\cdot}} (1-\tau)^{m_{\cdot \cdot}} &\notag
\end{align}
\normalsize
where $n_{dj}^k$ is a number of word tokens in the $d$th document with the $j$th word in the vocabulary assigned to the $k$th topic where $s_j = 1$, and $m_{d j}$ is a number of word tokens in $d$th document with the $j$th word where $s_j = 0$. The dots represent the marginal counts, so $m_{\cdot j}$ represents the number of word tokens of $j$th word across the corpus.

\begin{figure*}[t!]
  \centering
 
    \begin{tabular}{cccccccccccc}
    Model&1&2&3&4&5&6&7&8&9&10&NI\\
    \hline
Topics	&\includegraphics[width=0.035\linewidth]{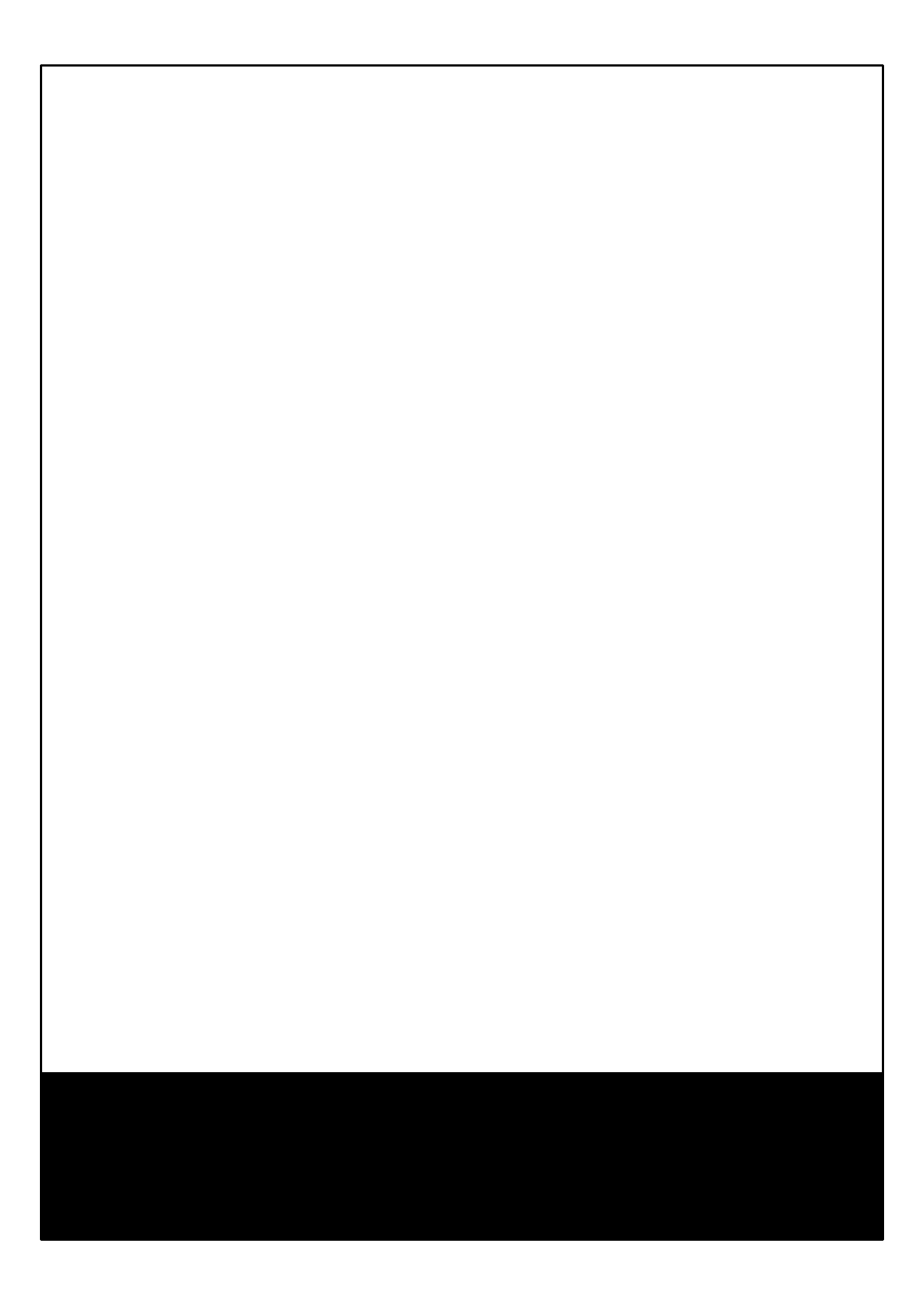}&\includegraphics[width=0.035\linewidth]{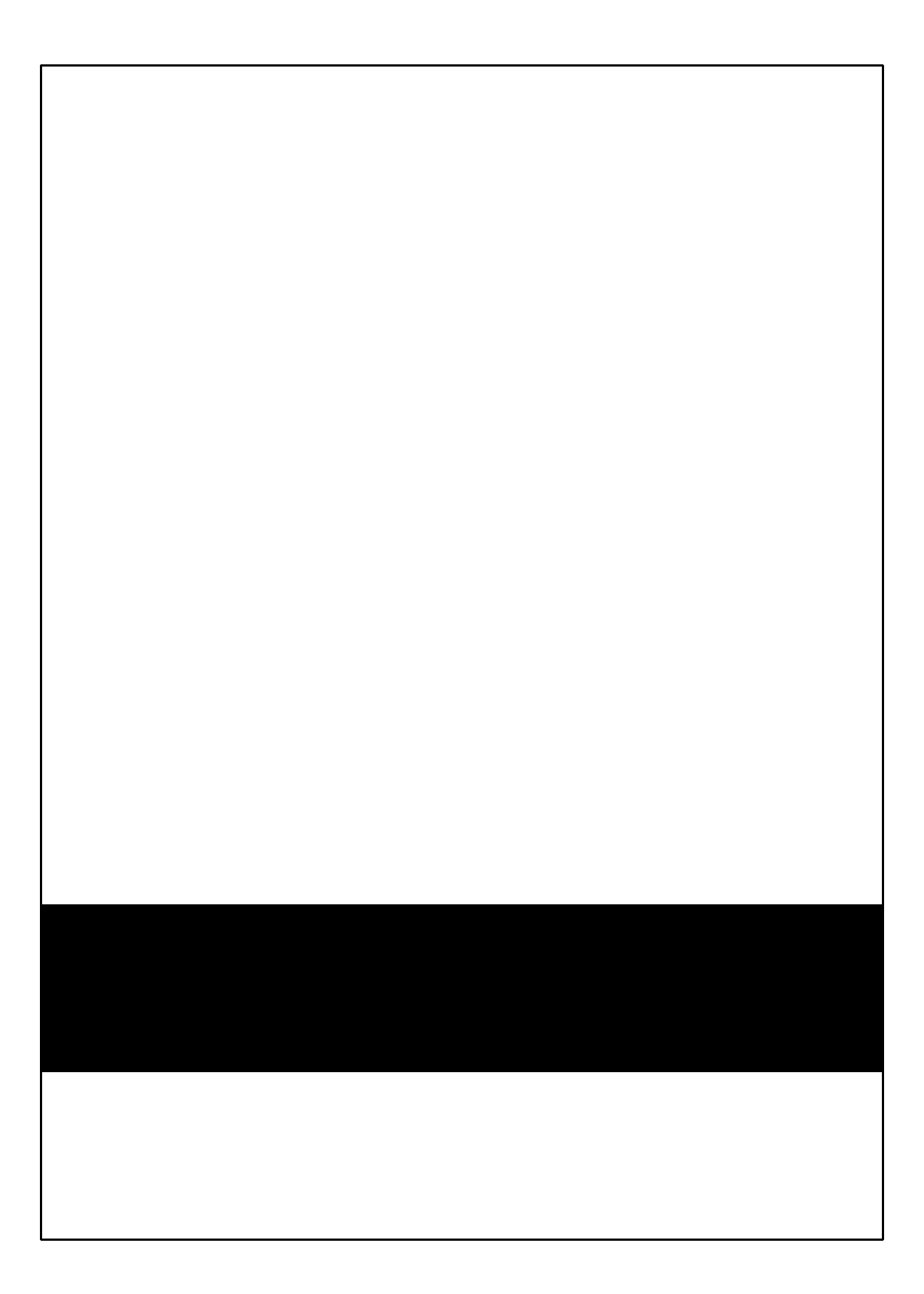}&\includegraphics[width=0.037\linewidth]{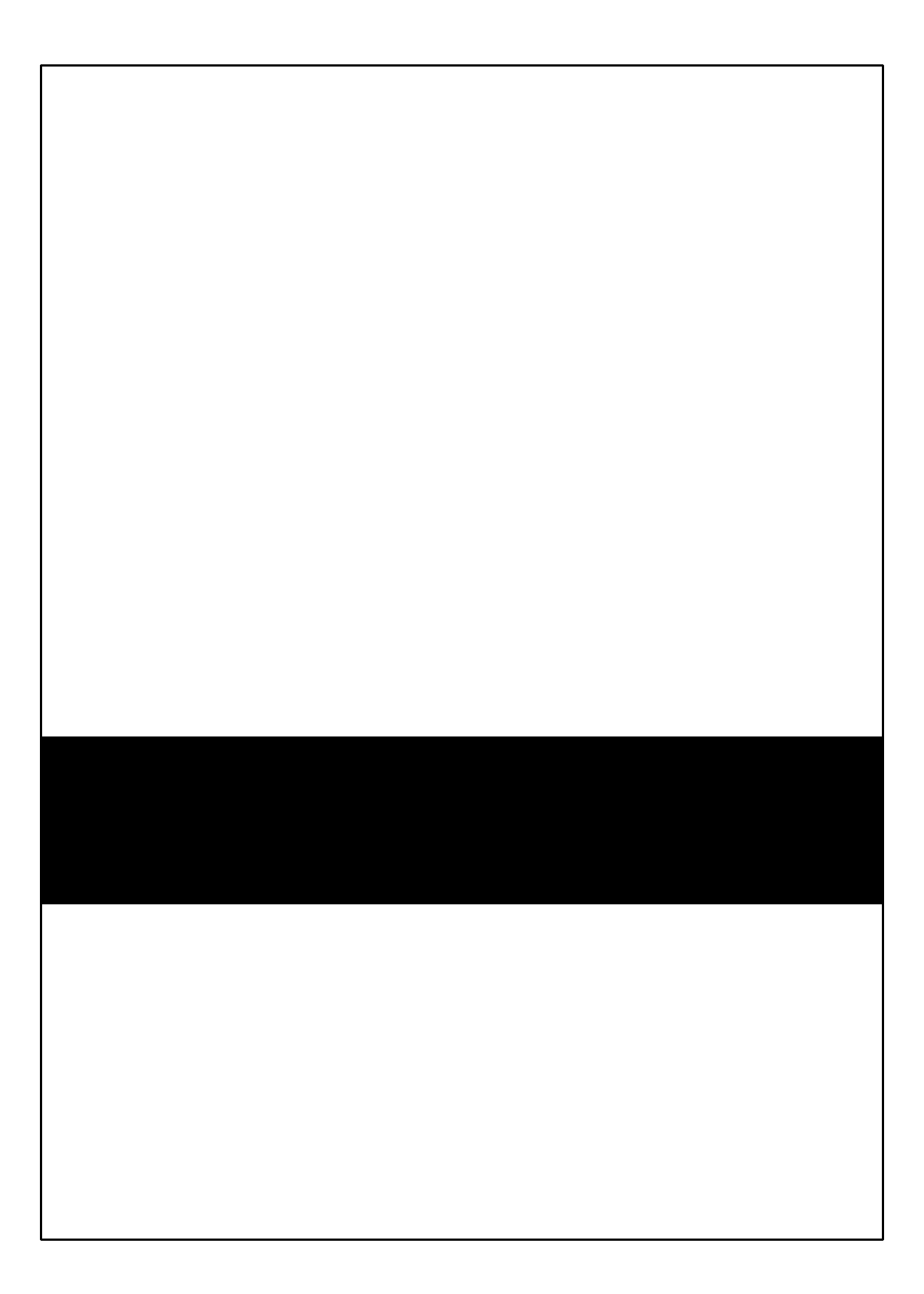}&\includegraphics[width=0.037\linewidth]{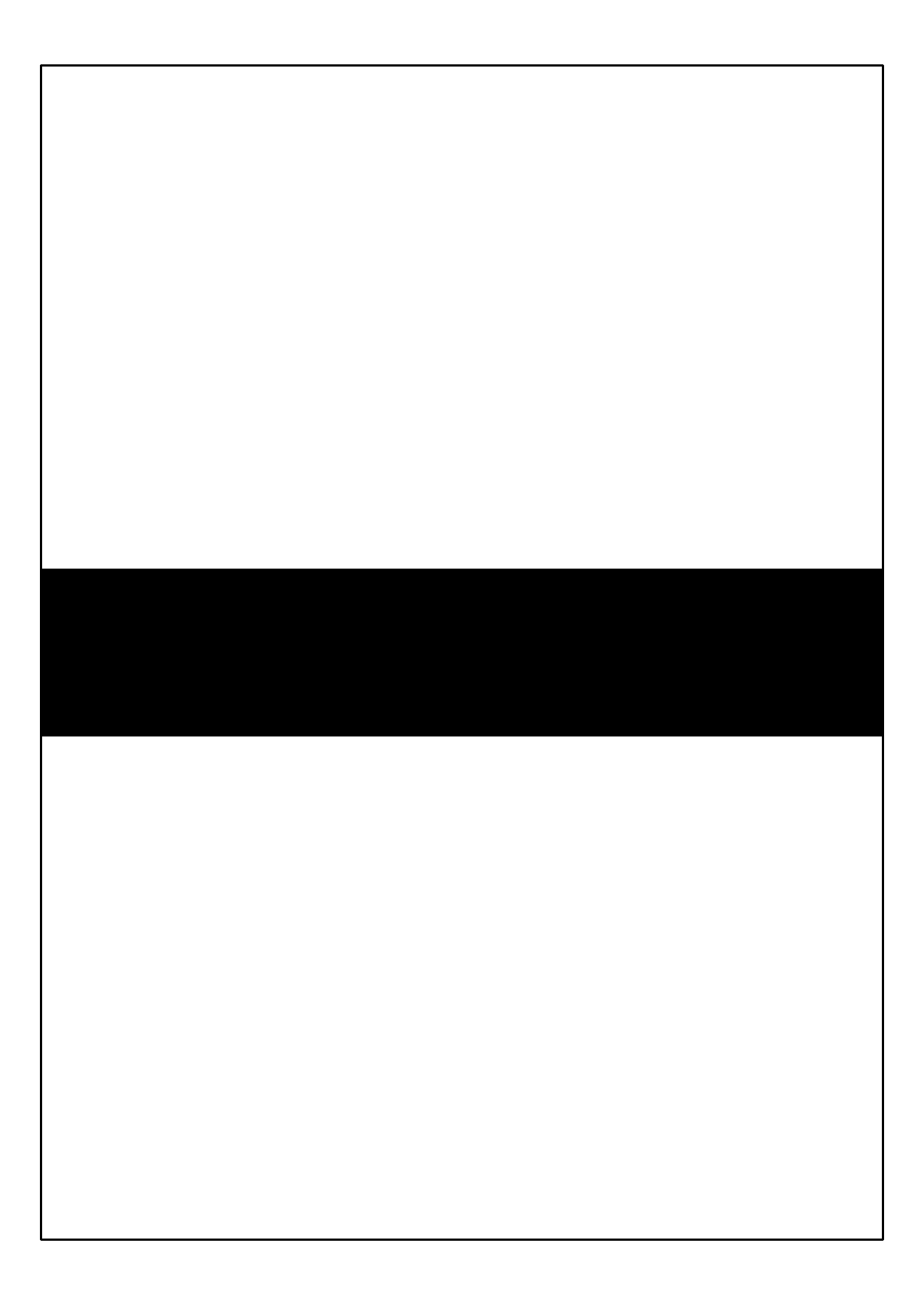}&\includegraphics[width=0.037\linewidth]{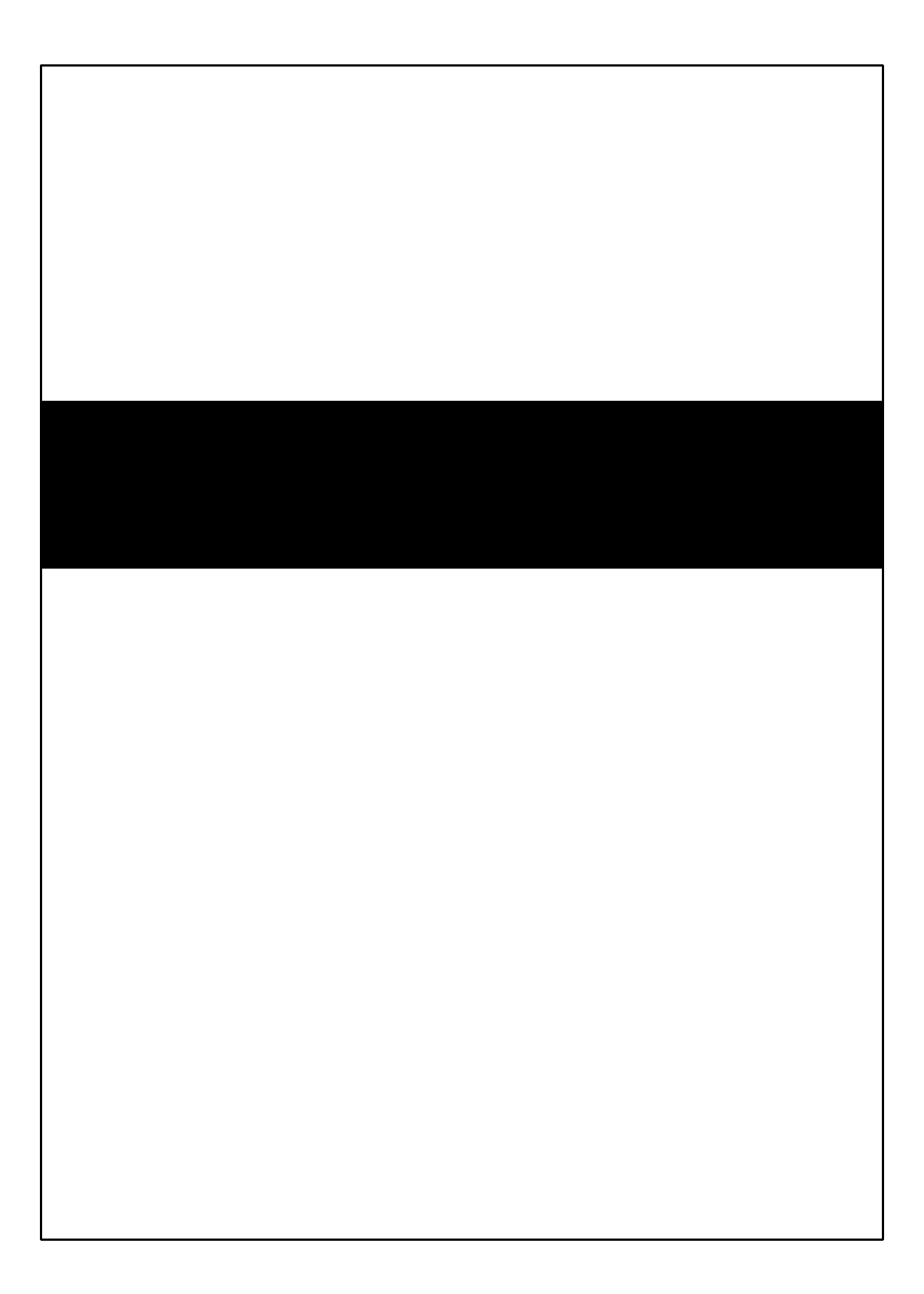}&\includegraphics[width=0.037\linewidth]{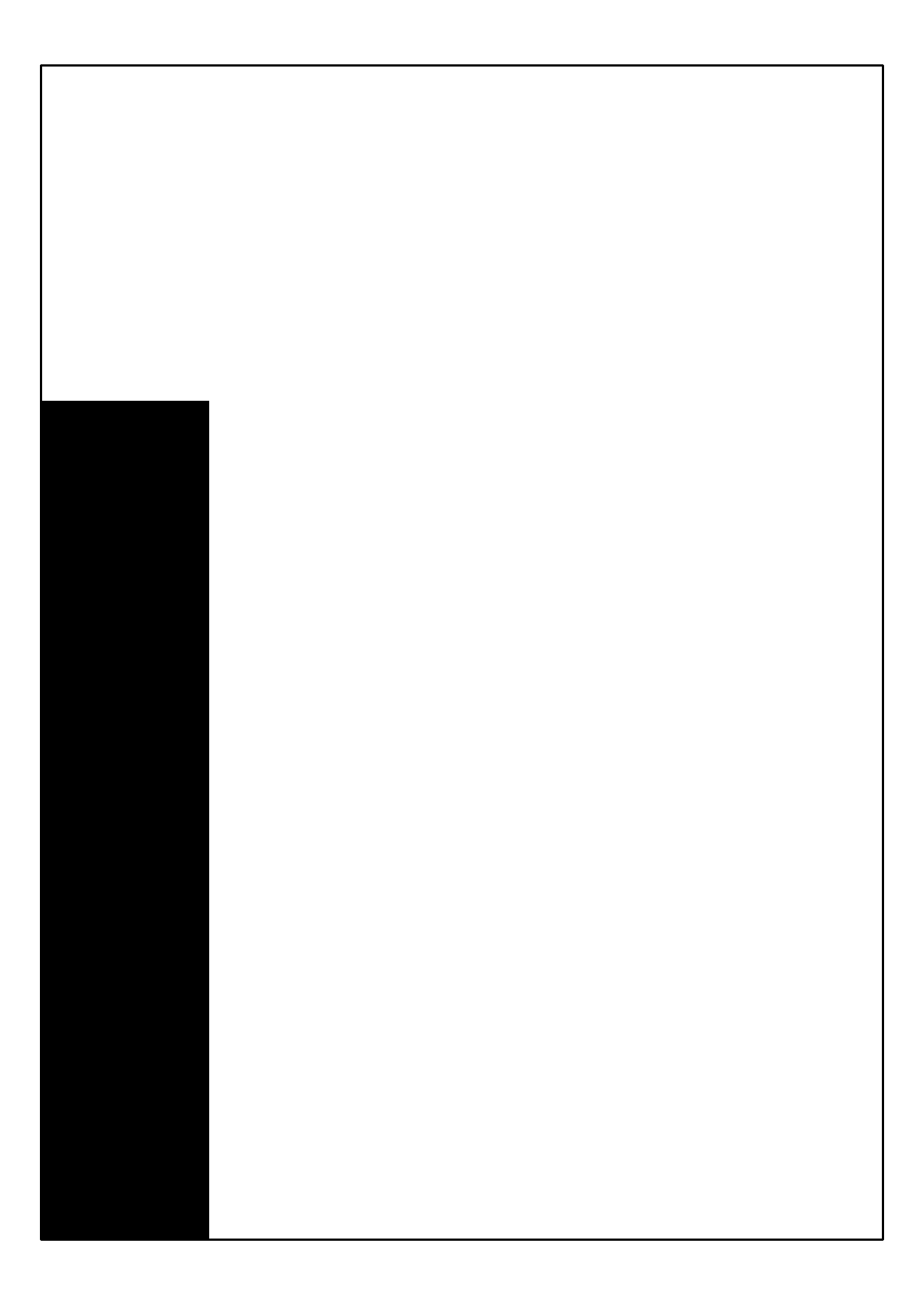}&\includegraphics[width=0.037\linewidth]{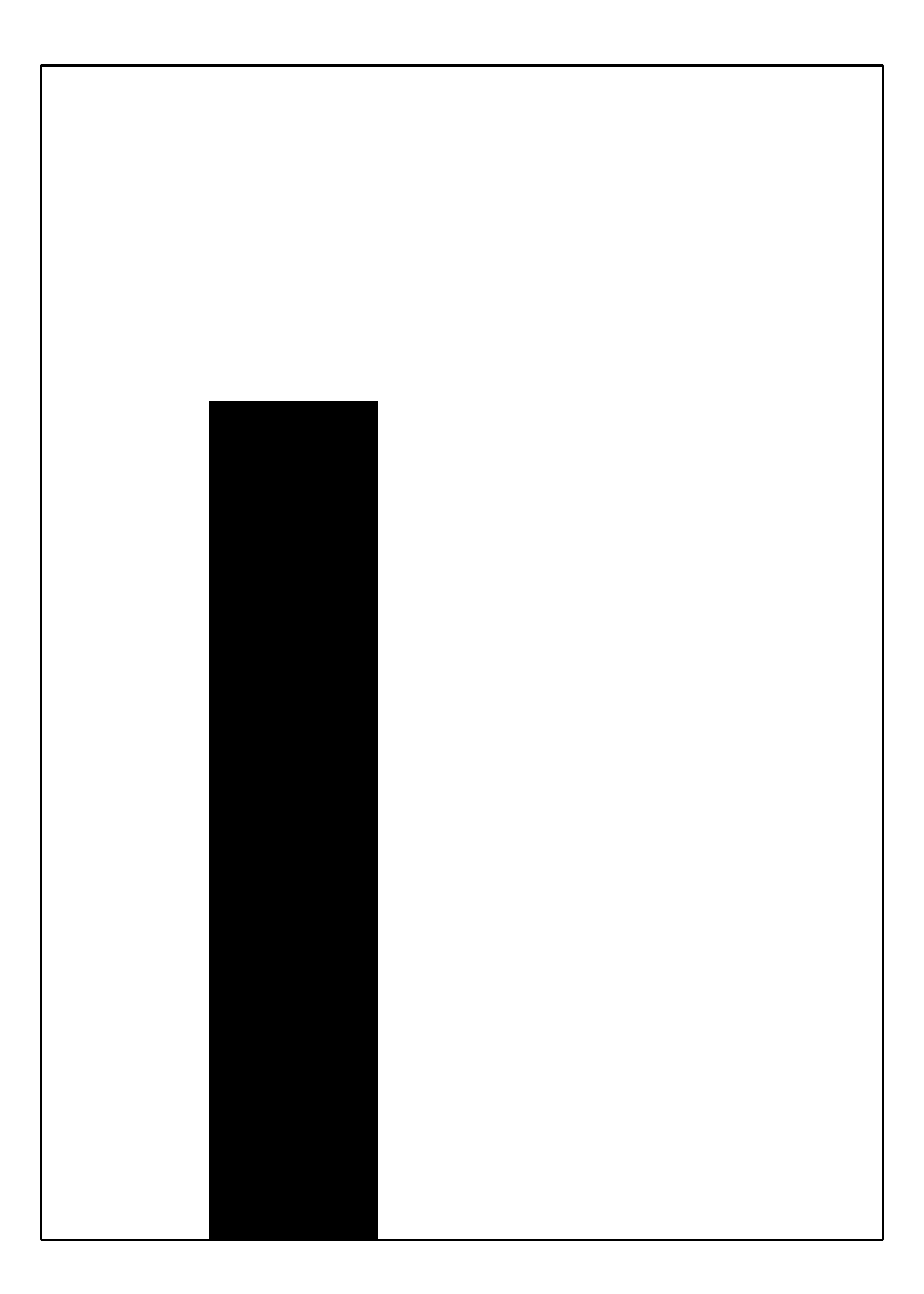}&\includegraphics[width=0.037\linewidth]{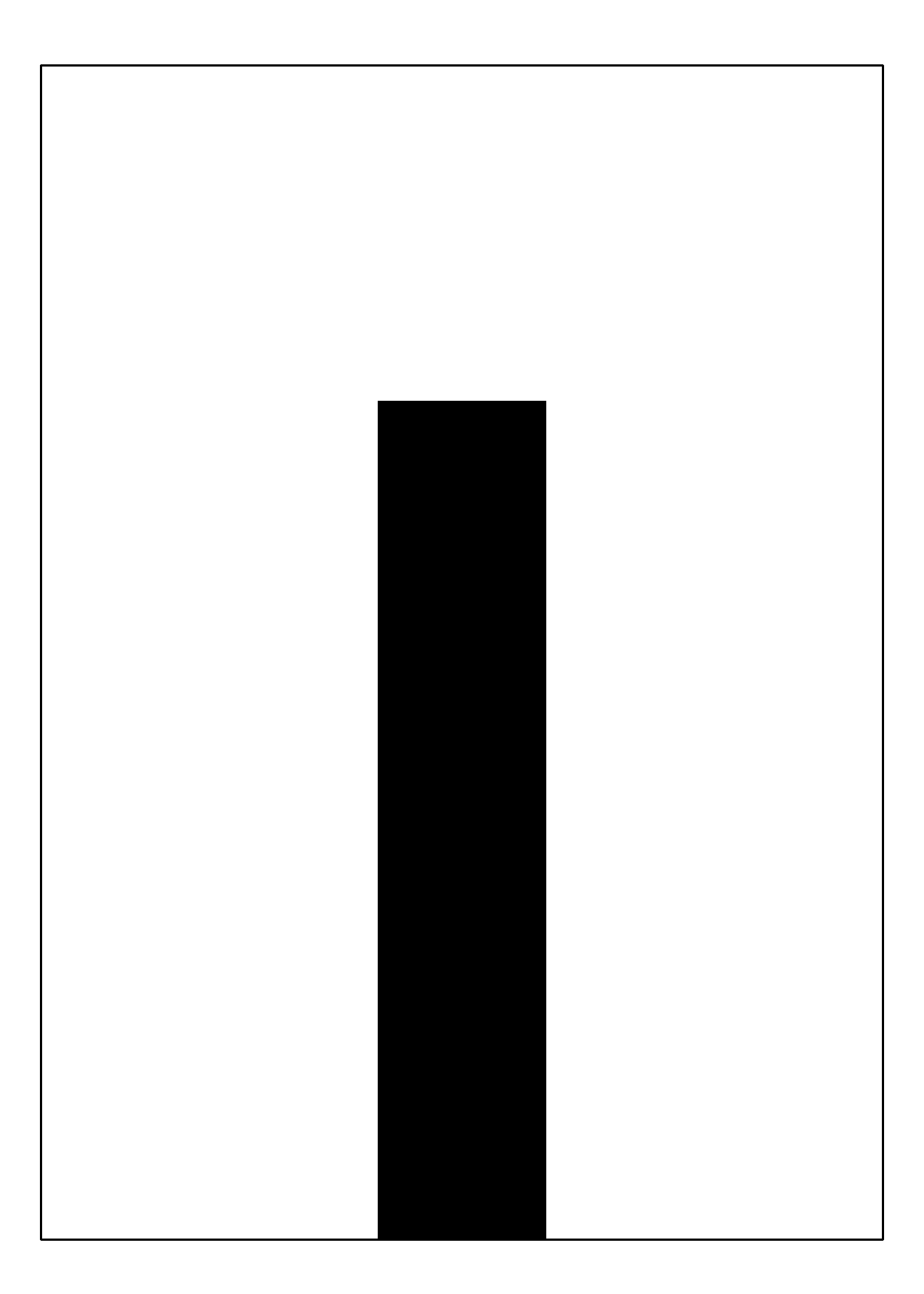}&\includegraphics[width=0.037\linewidth]{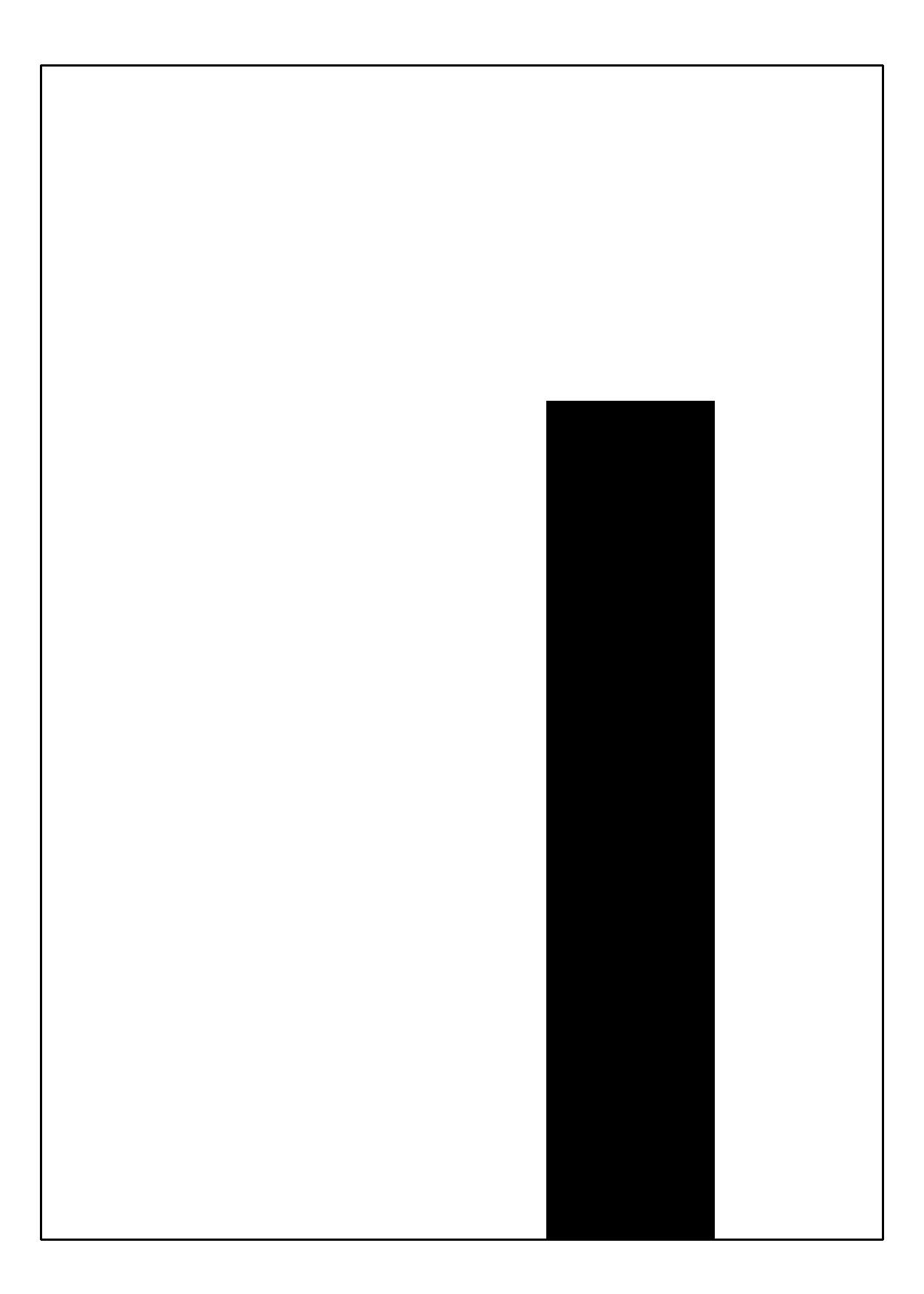}&\includegraphics[width=0.037\linewidth]{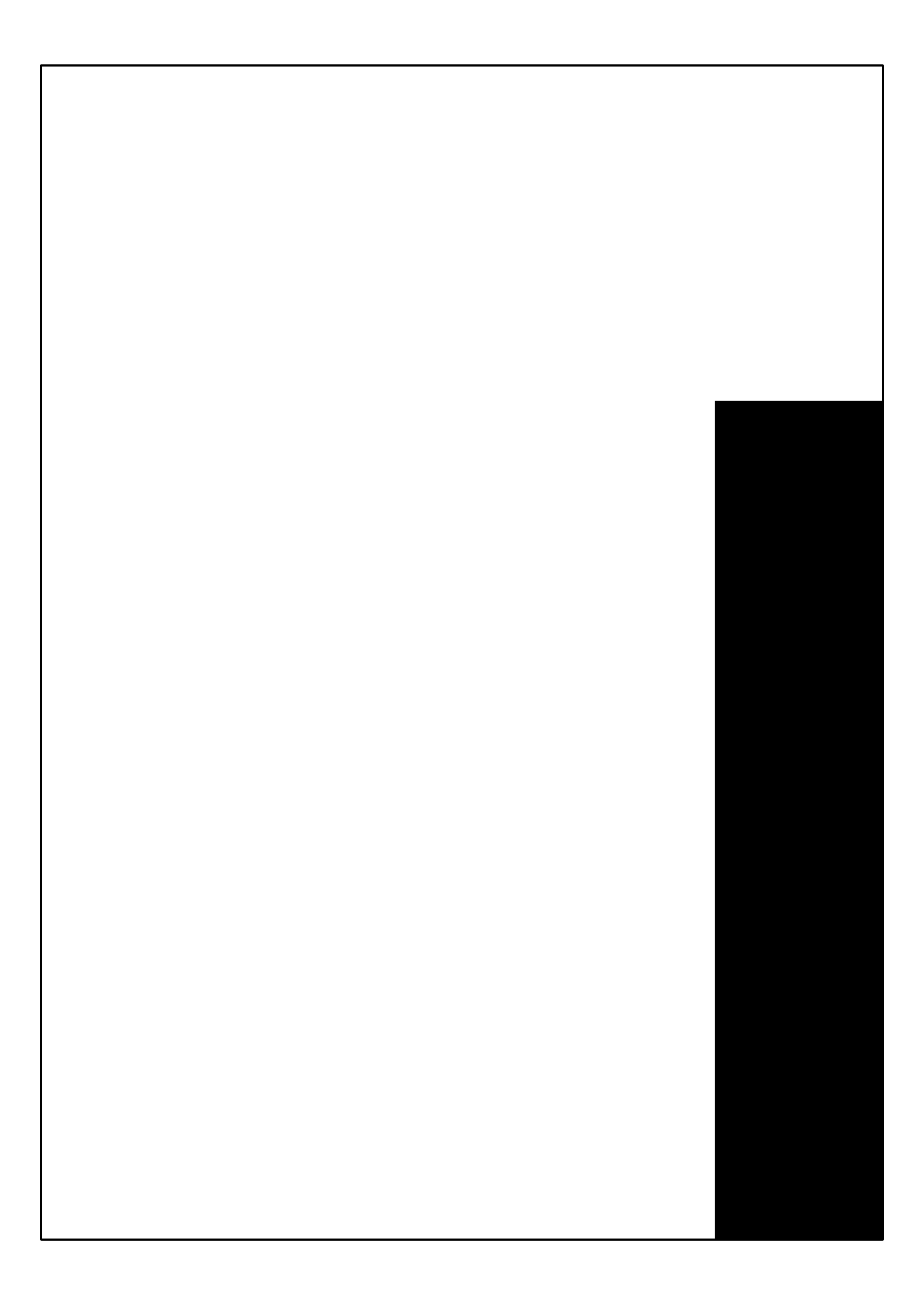}&\includegraphics[width=0.037\linewidth]{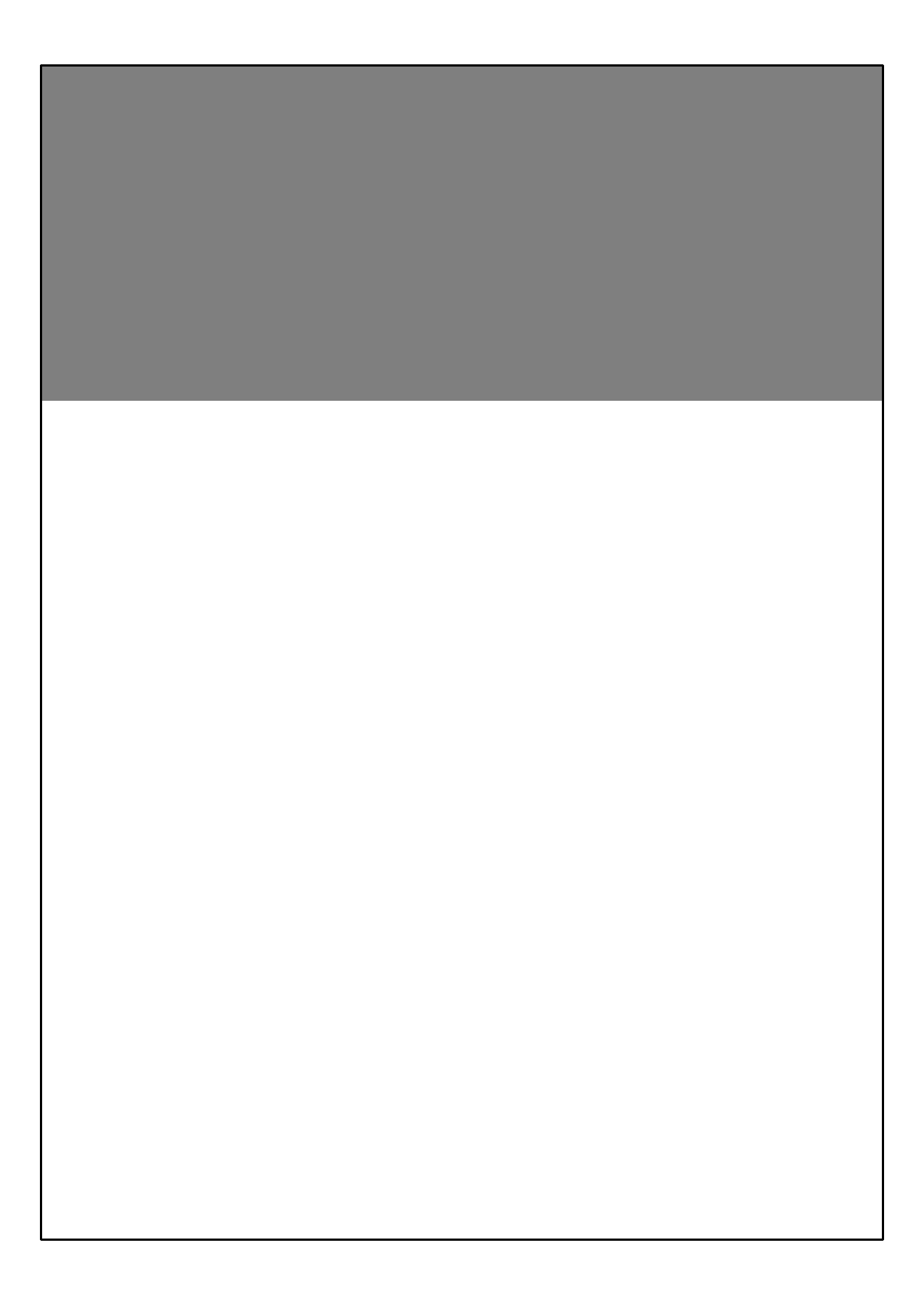}\\
vsLDA	&\includegraphics[width=0.037\linewidth]{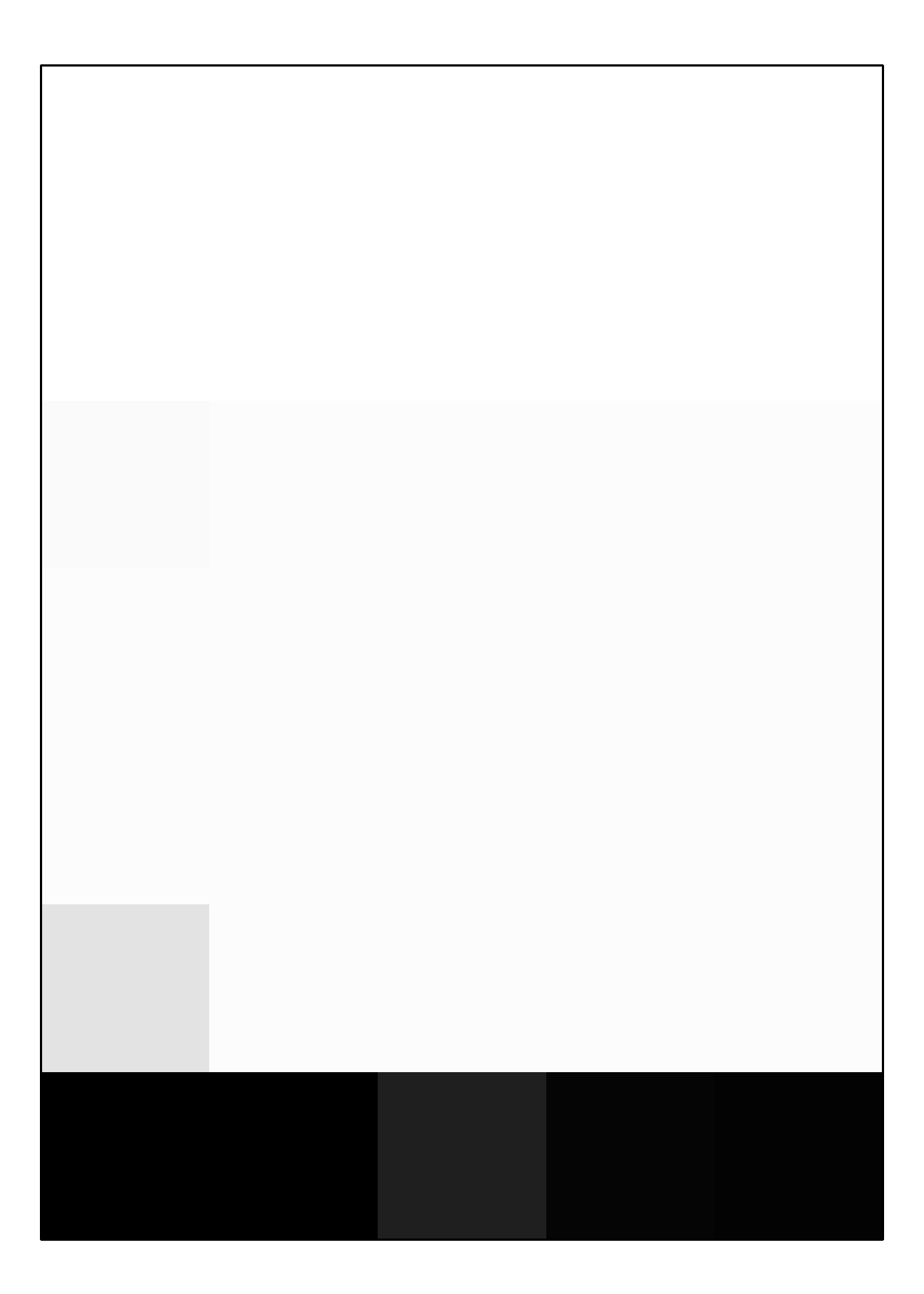}&\includegraphics[width=0.037\linewidth]{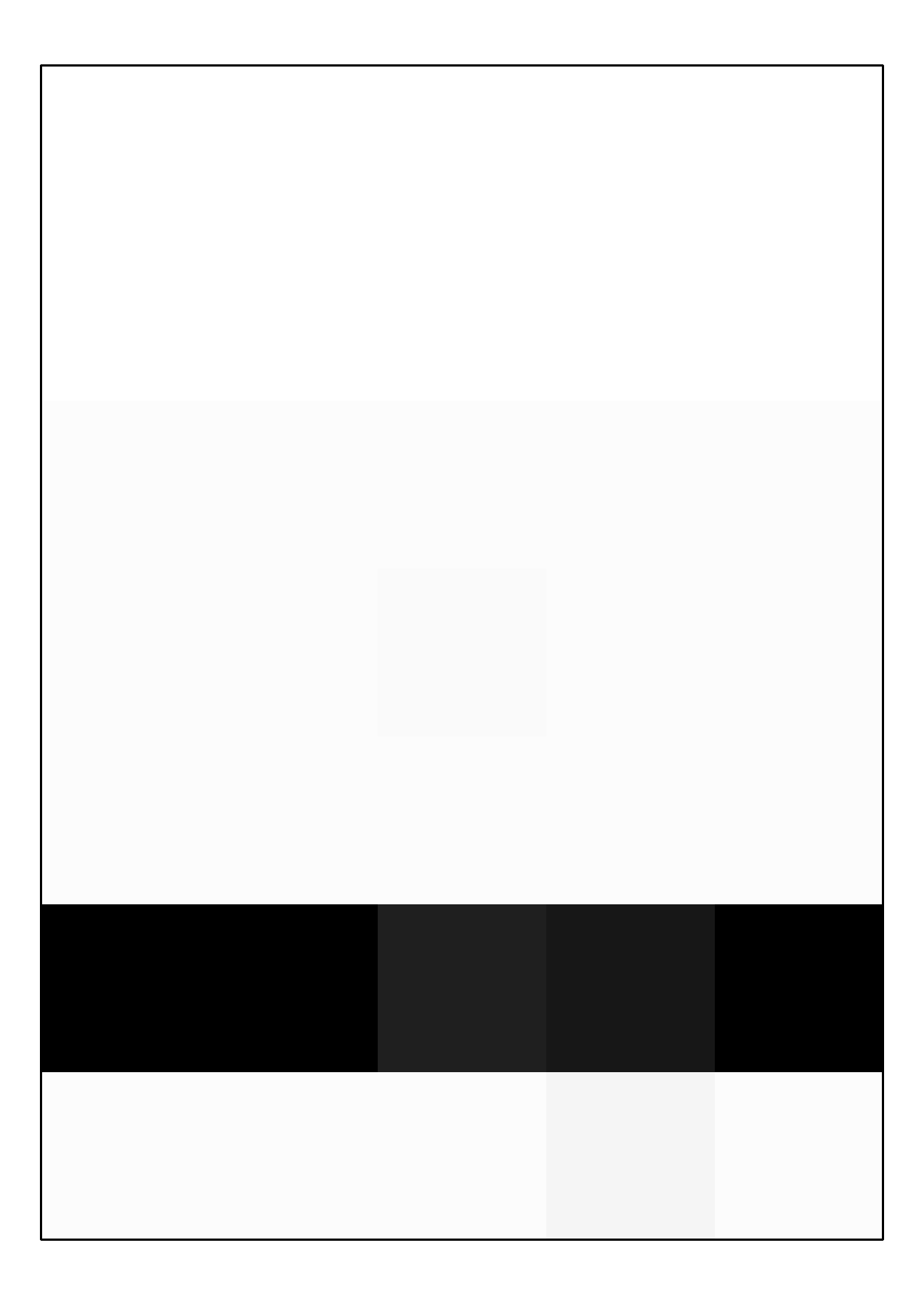}&\includegraphics[width=0.037\linewidth]{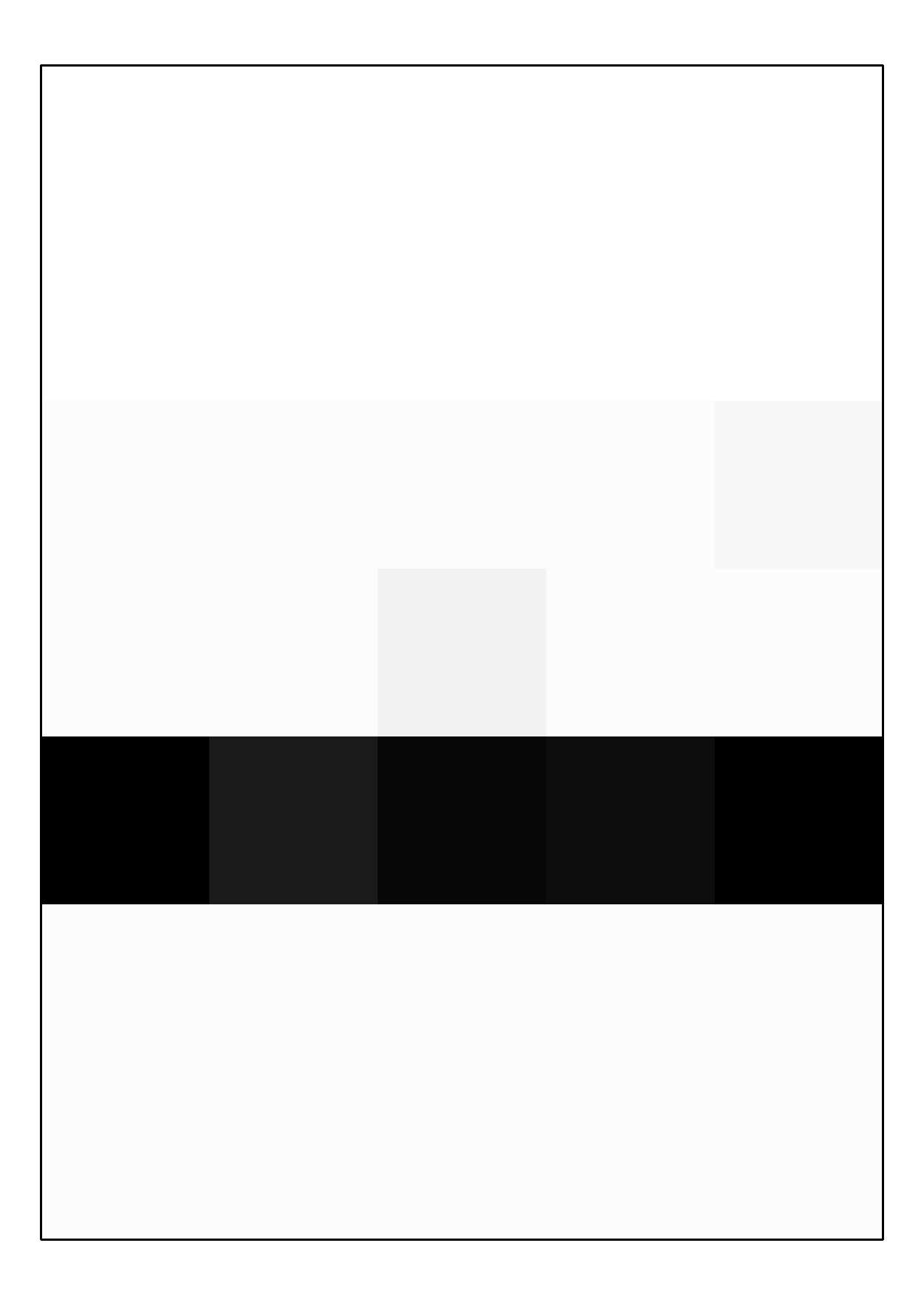}&\includegraphics[width=0.037\linewidth]{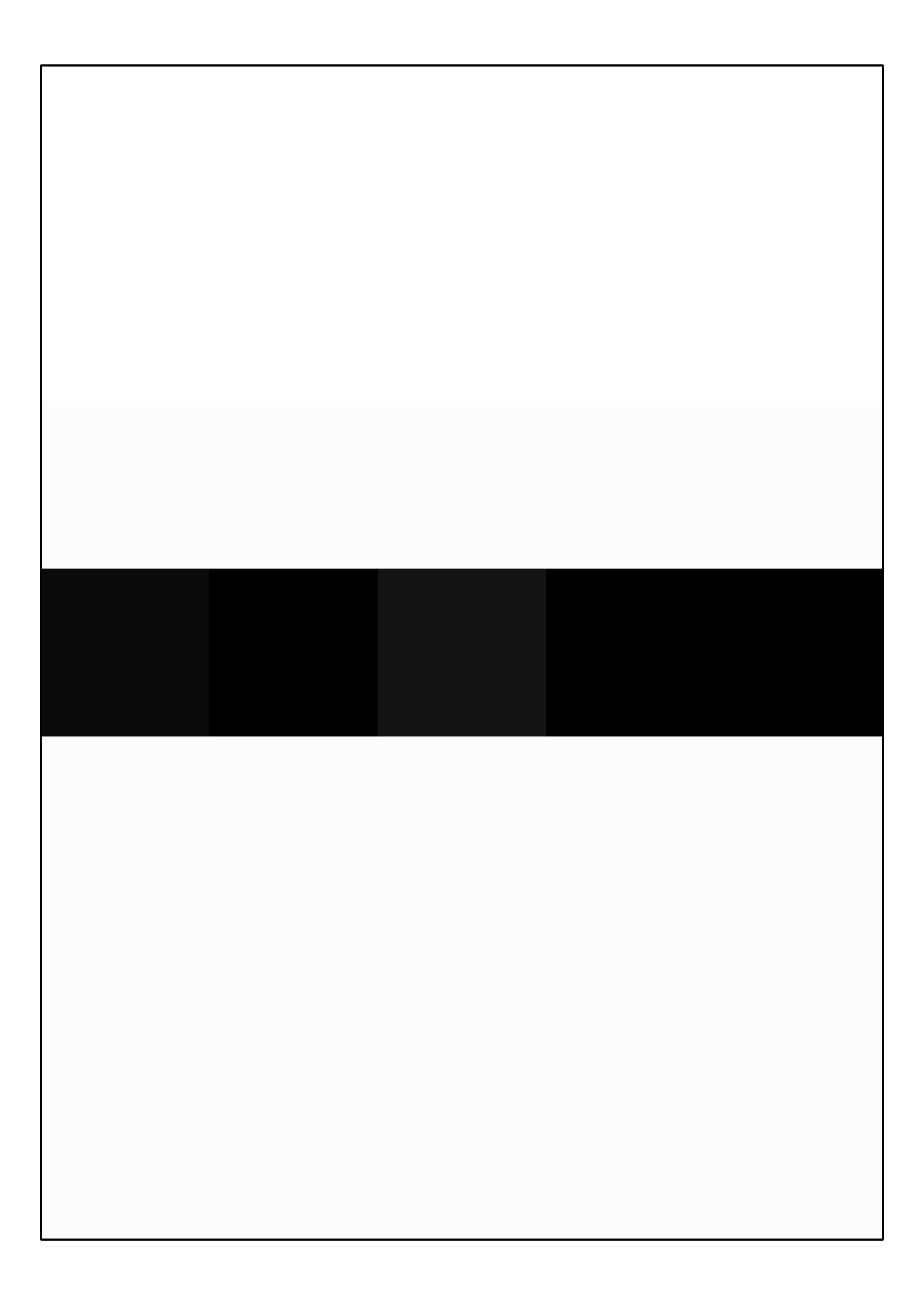}&\includegraphics[width=0.037\linewidth]{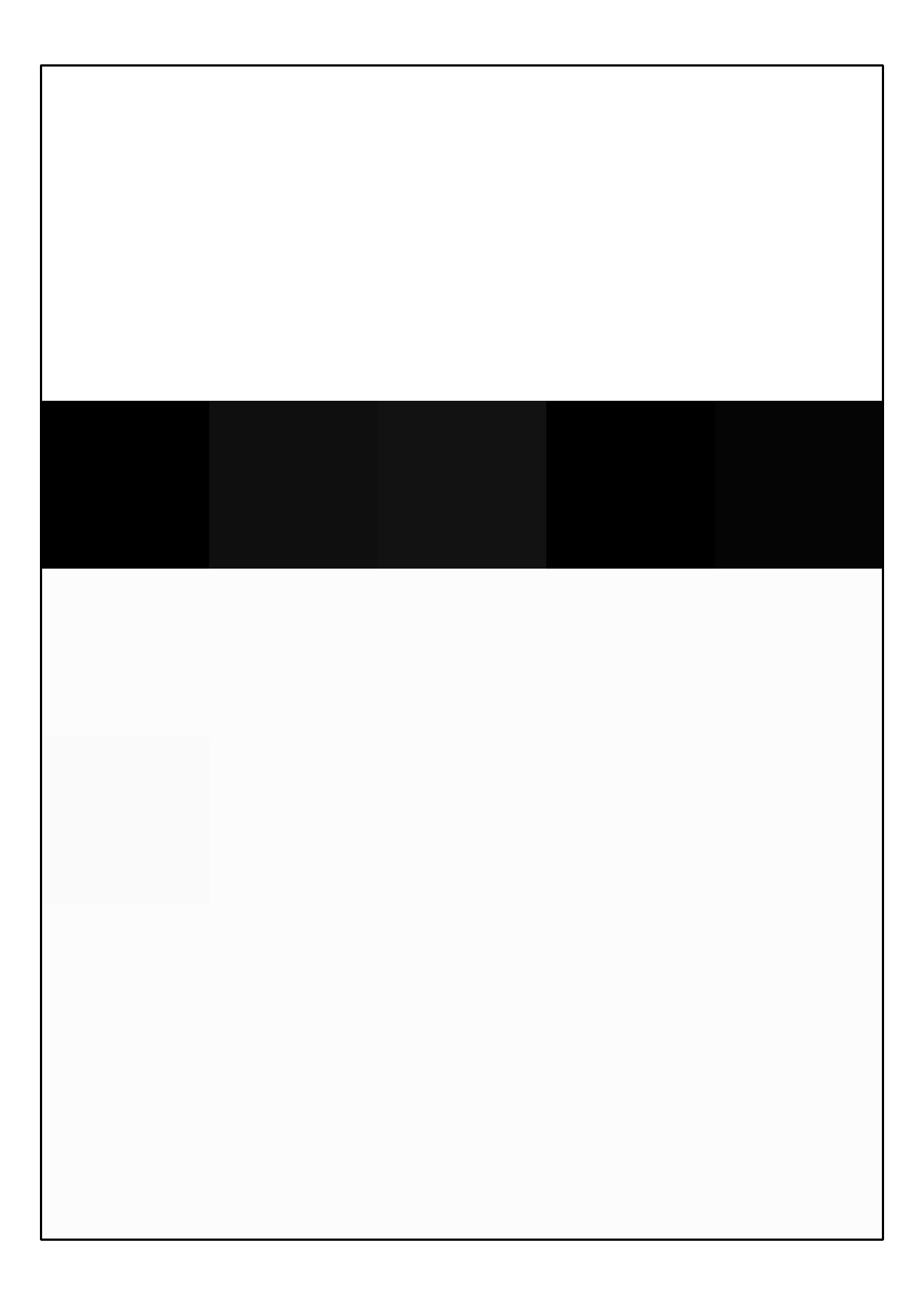}&\includegraphics[width=0.037\linewidth]{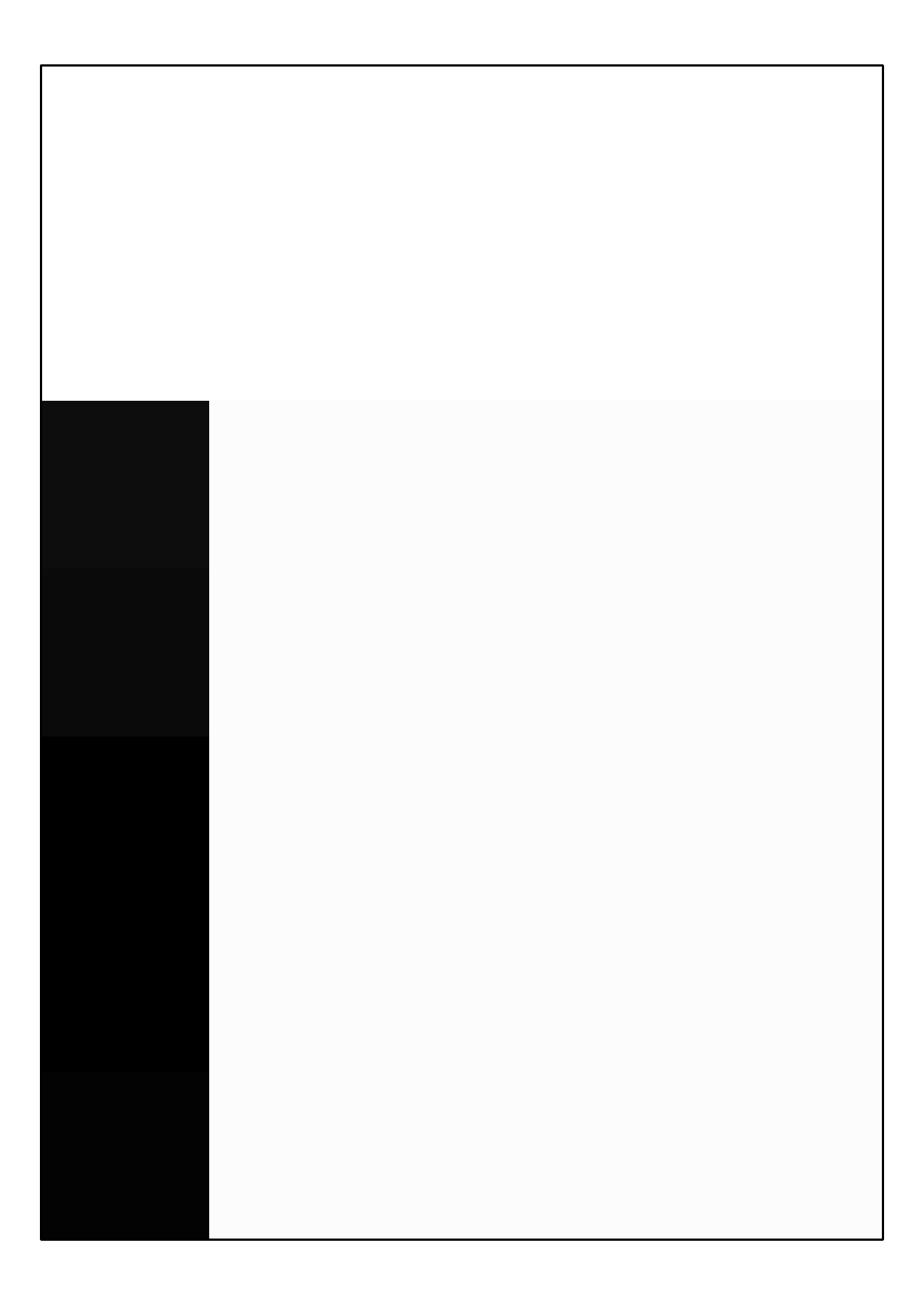}&\includegraphics[width=0.037\linewidth]{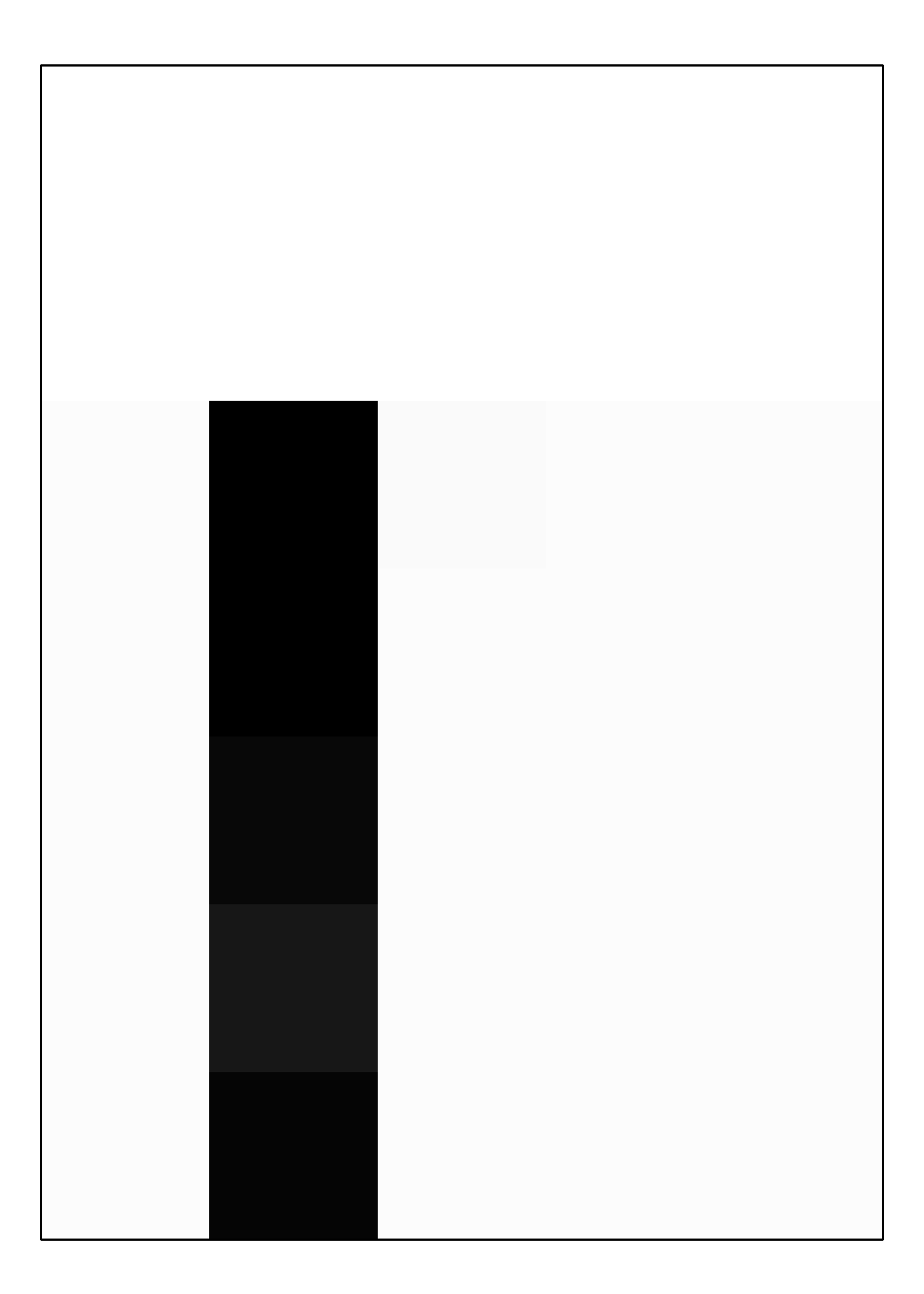}&\includegraphics[width=0.037\linewidth]{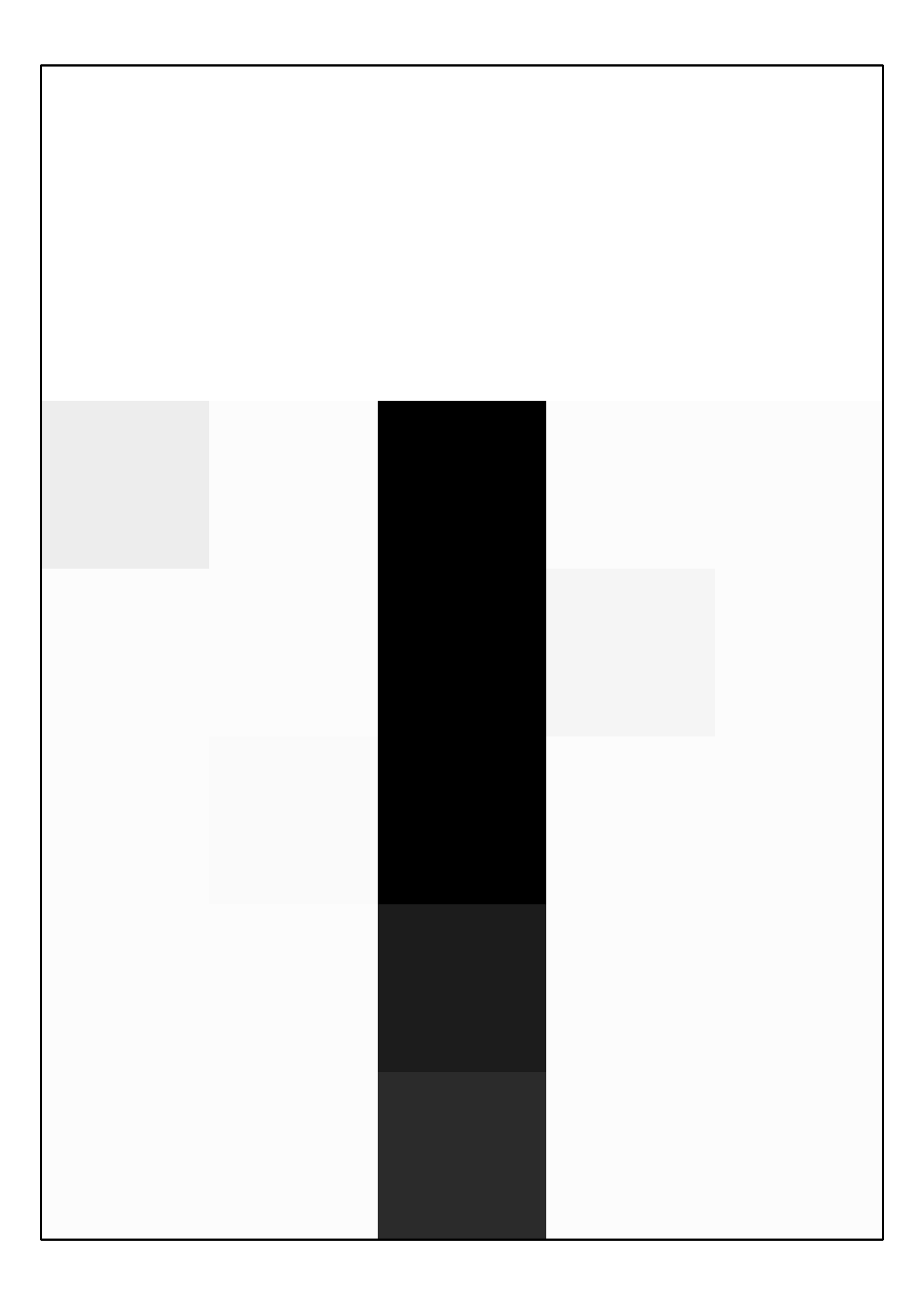}&\includegraphics[width=0.037\linewidth]{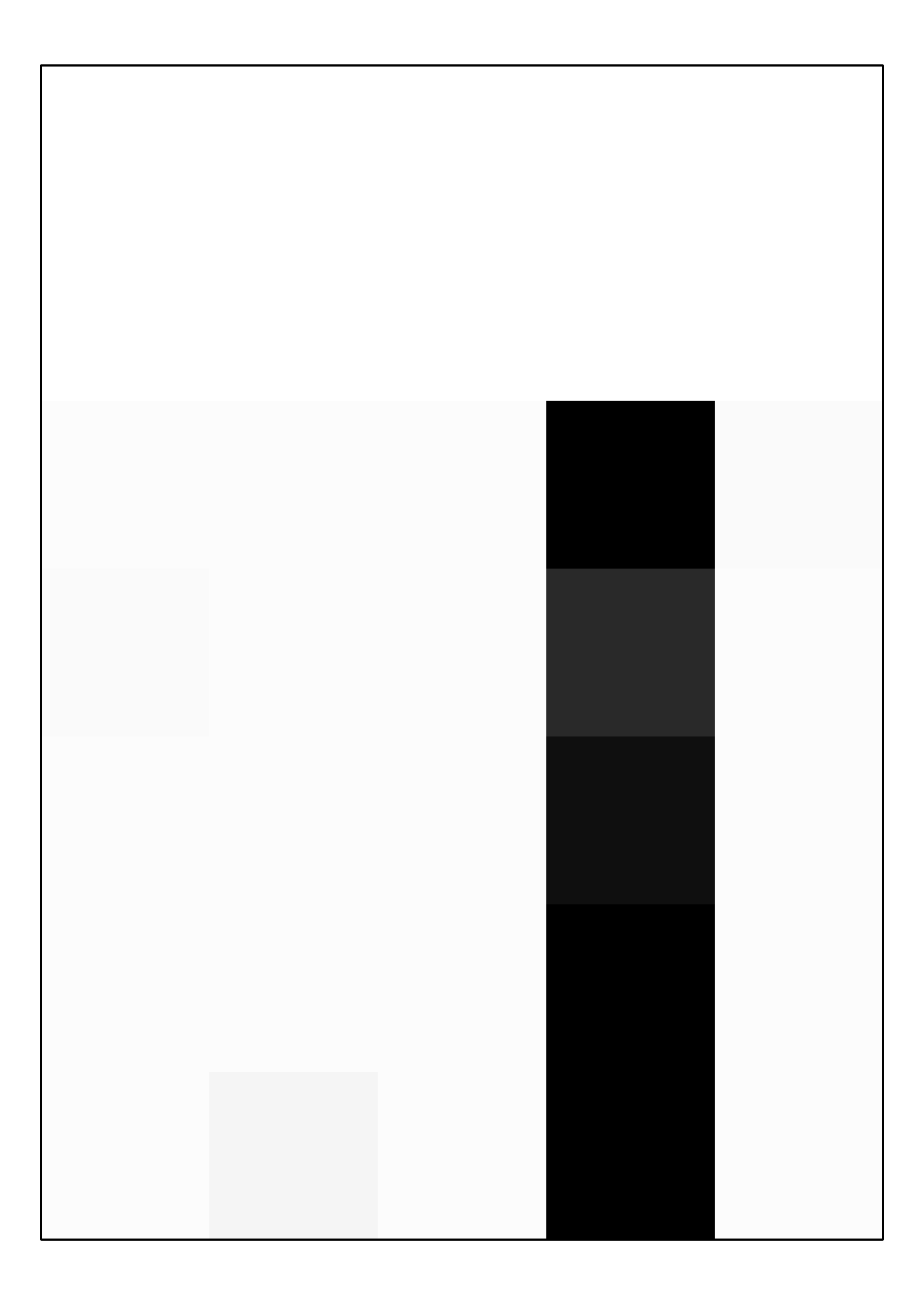}&\includegraphics[width=0.037\linewidth]{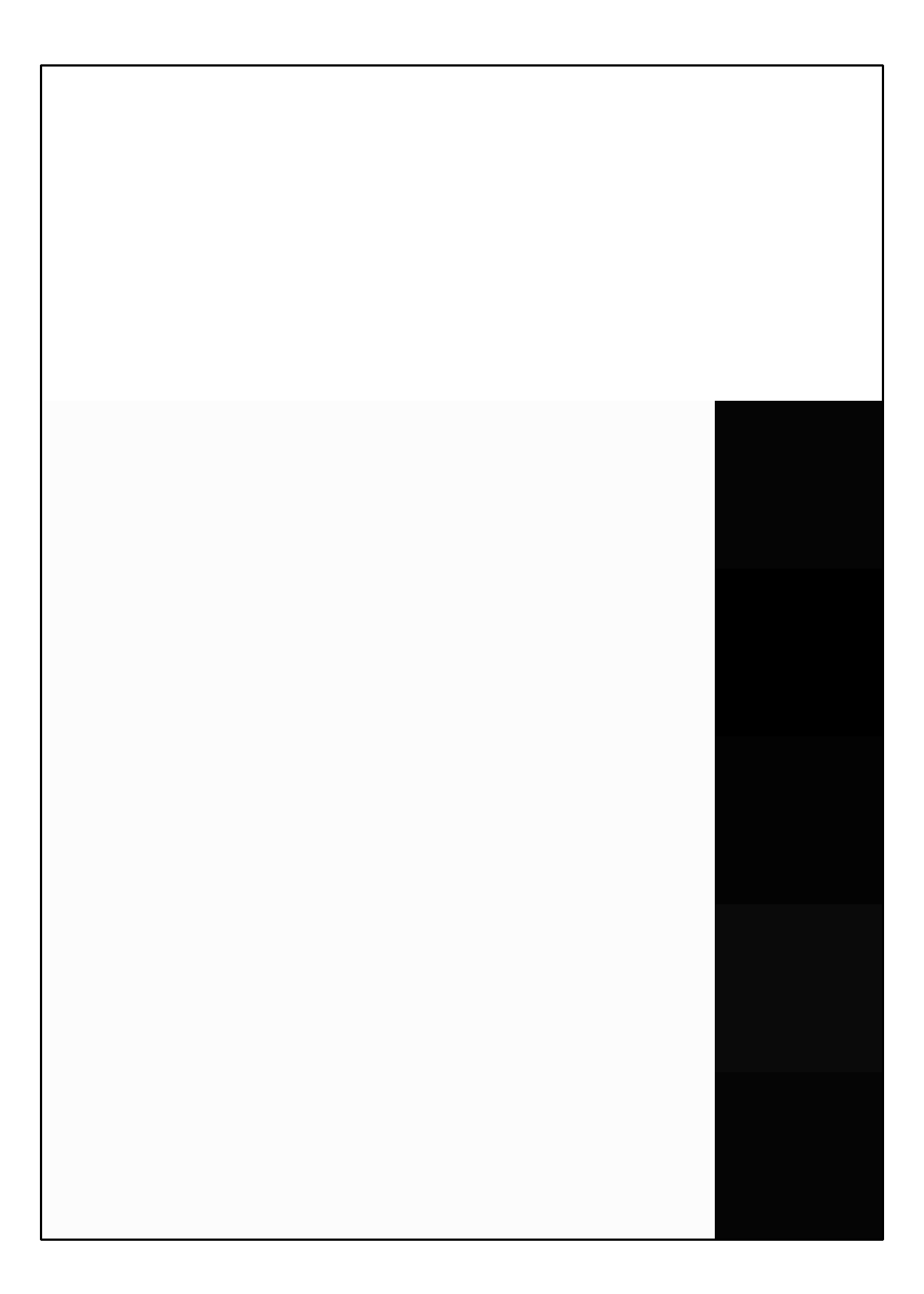}&\includegraphics[width=0.037\linewidth]{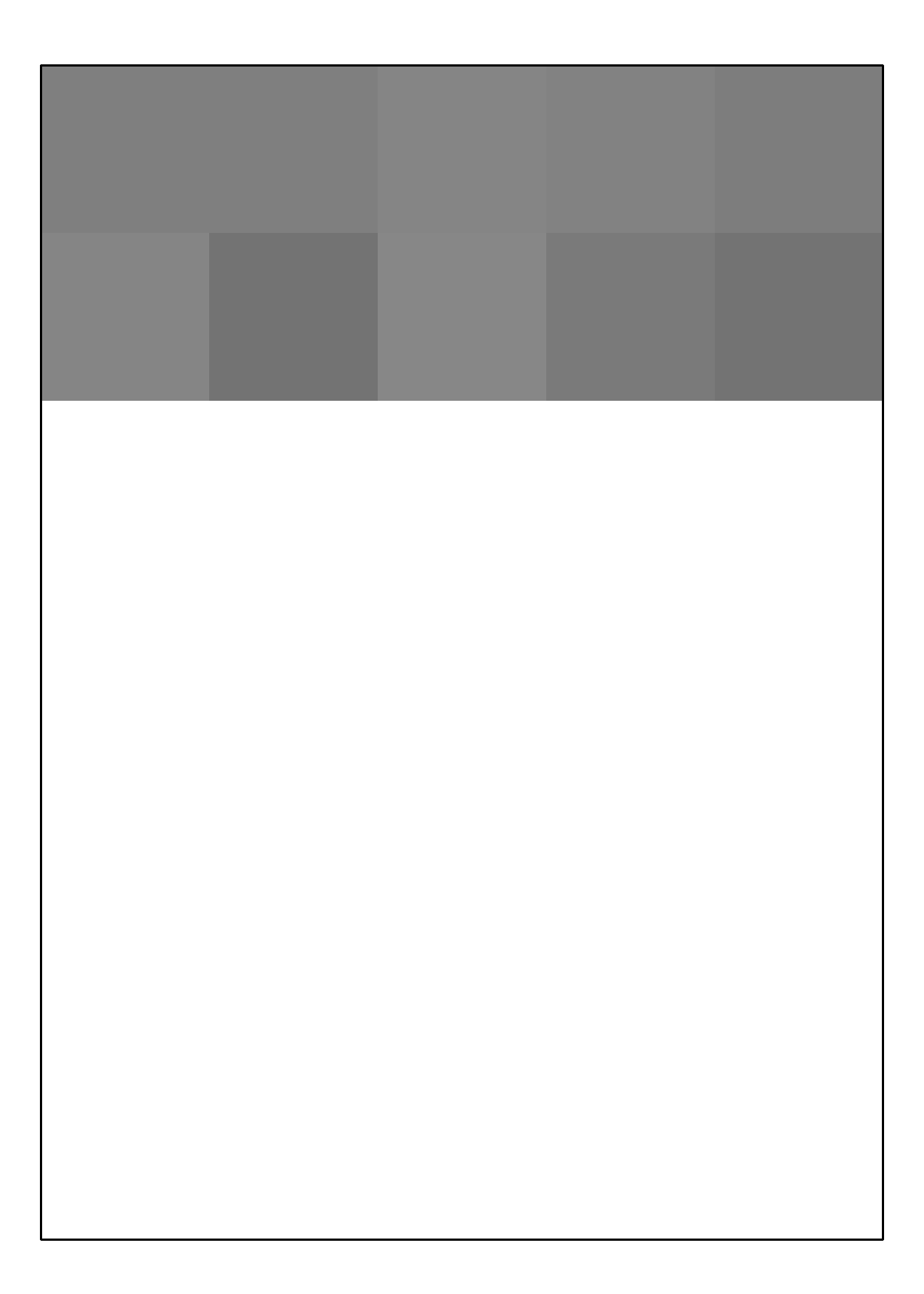}\\
LDA10	&\includegraphics[width=0.037\linewidth]{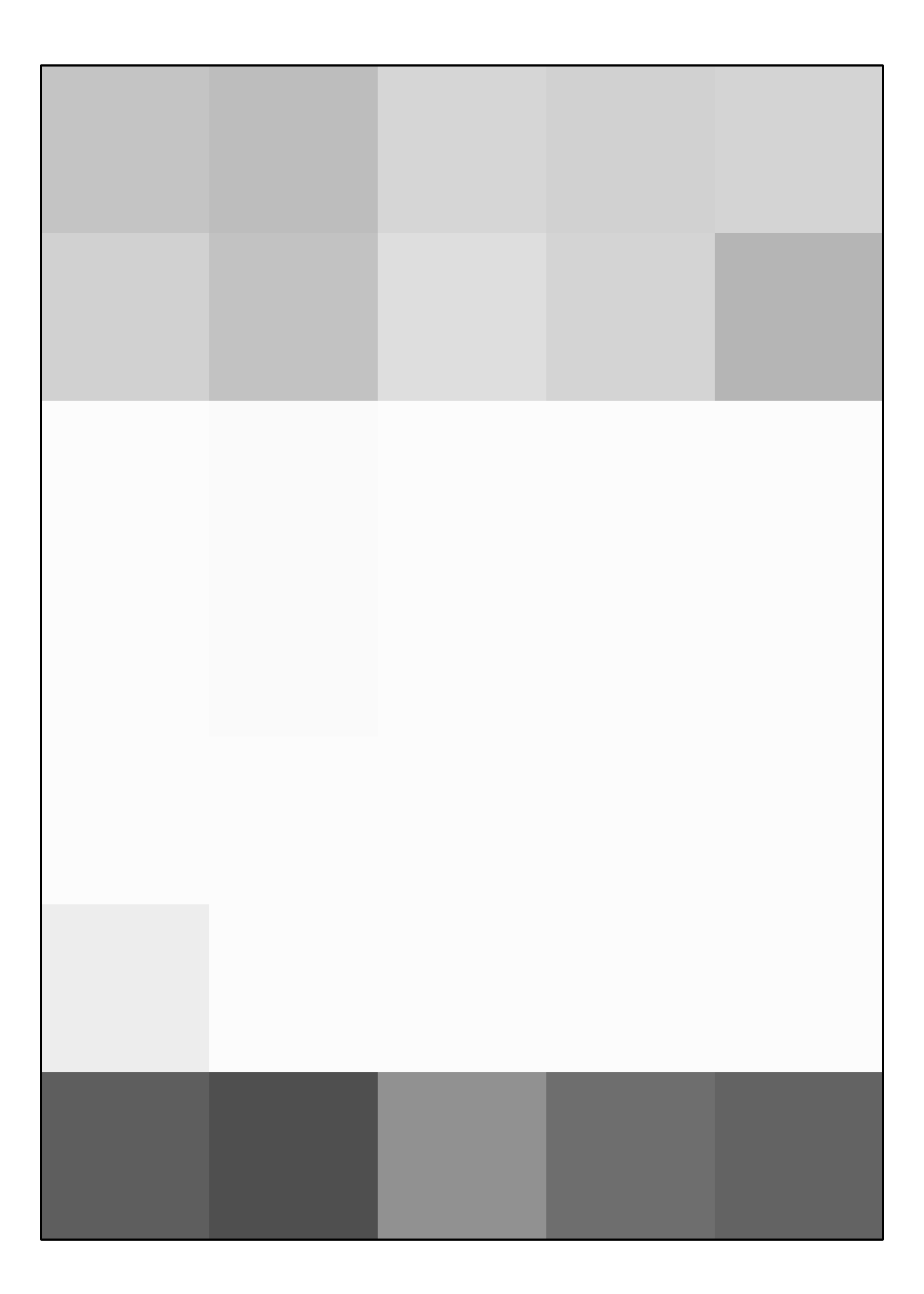}&\includegraphics[width=0.037\linewidth]{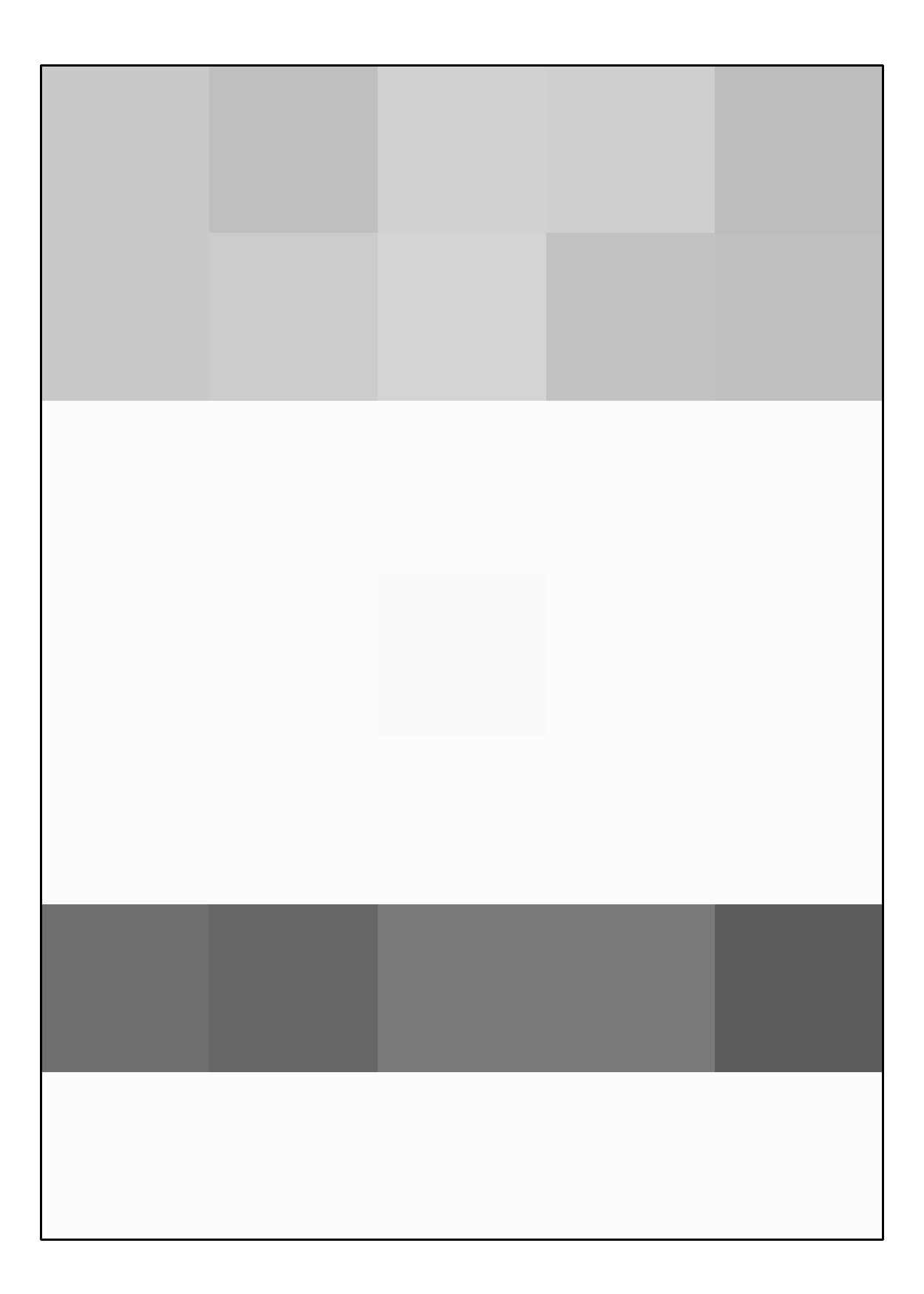}&\includegraphics[width=0.037\linewidth]{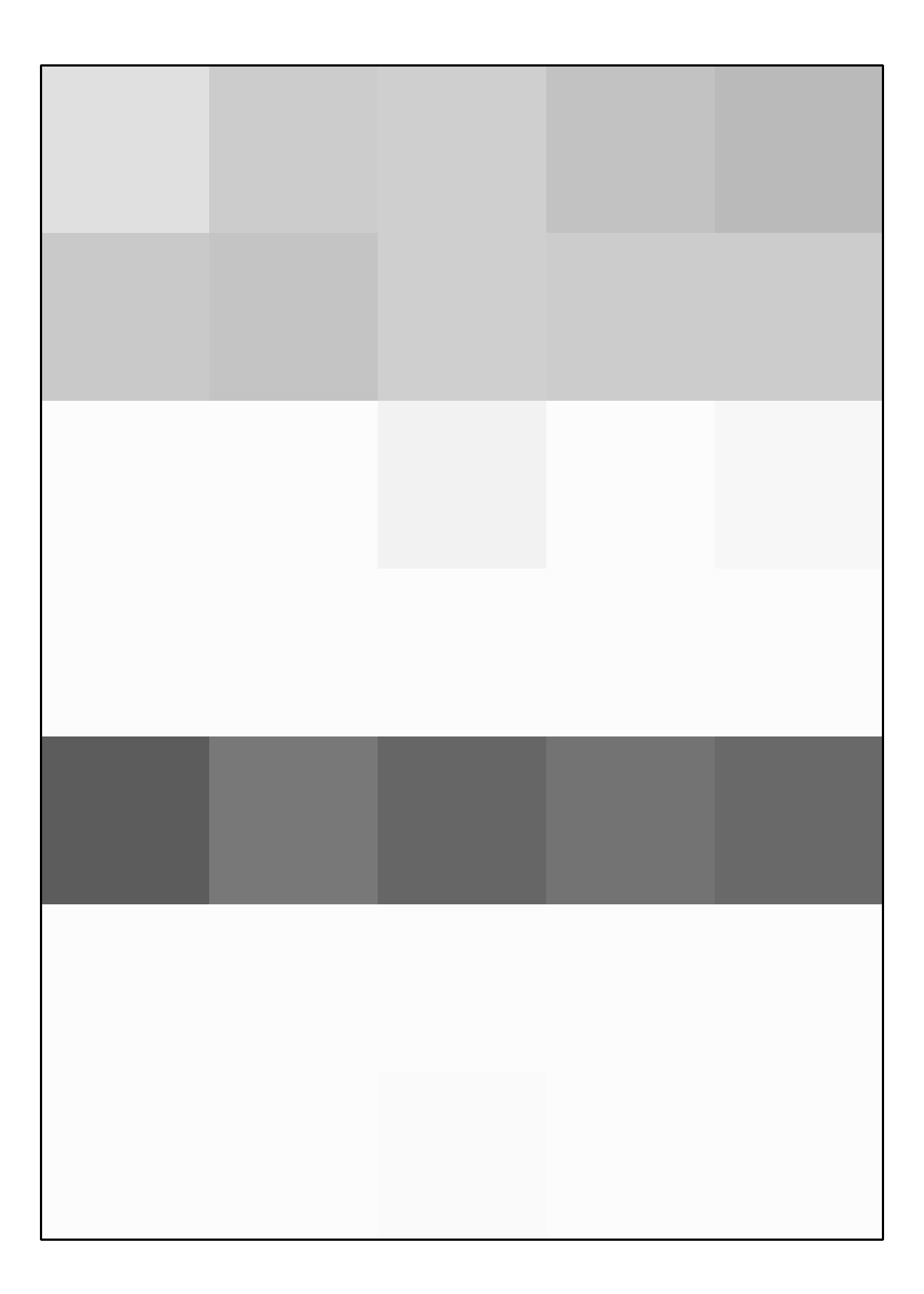}&\includegraphics[width=0.037\linewidth]{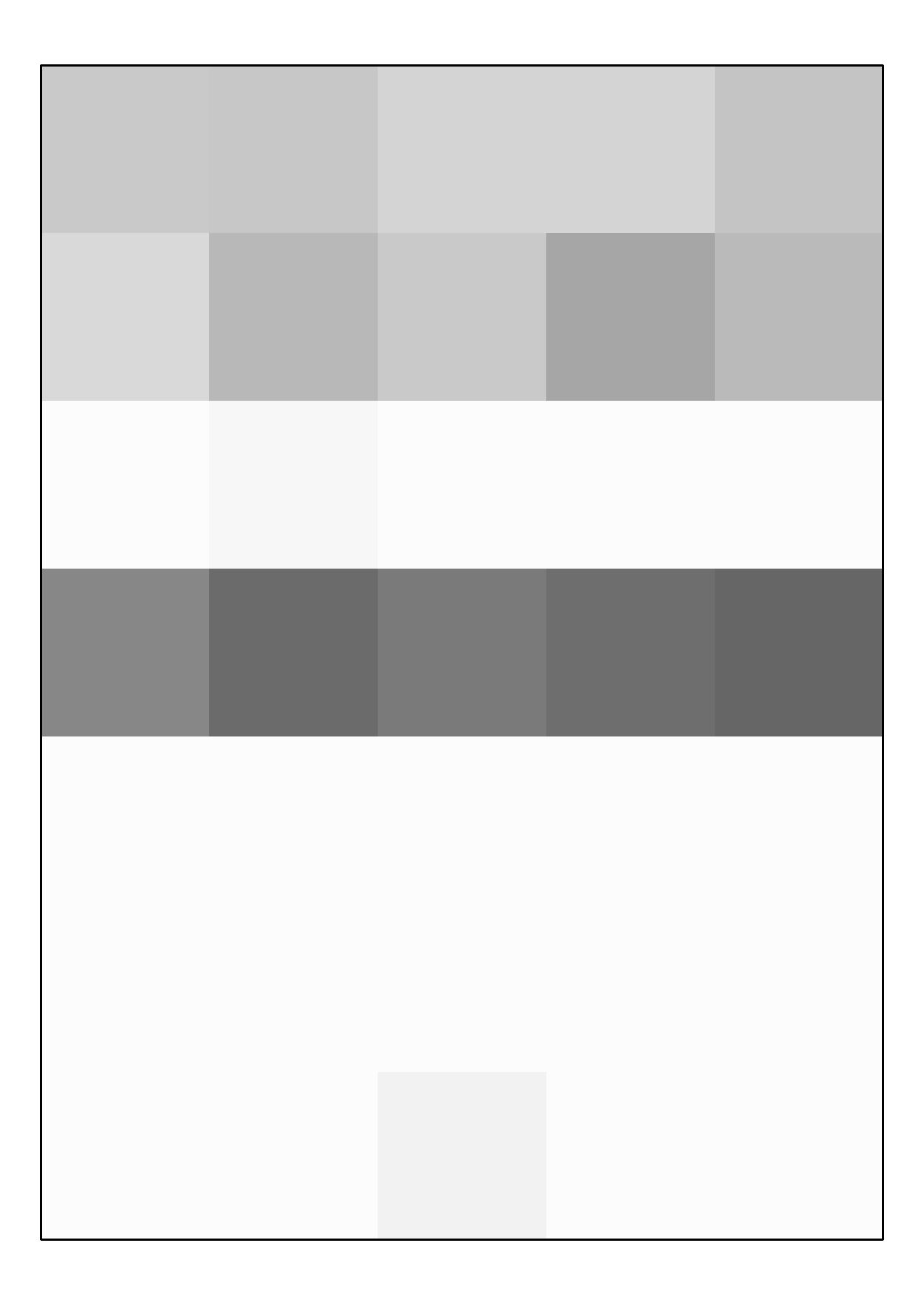}&\includegraphics[width=0.037\linewidth]{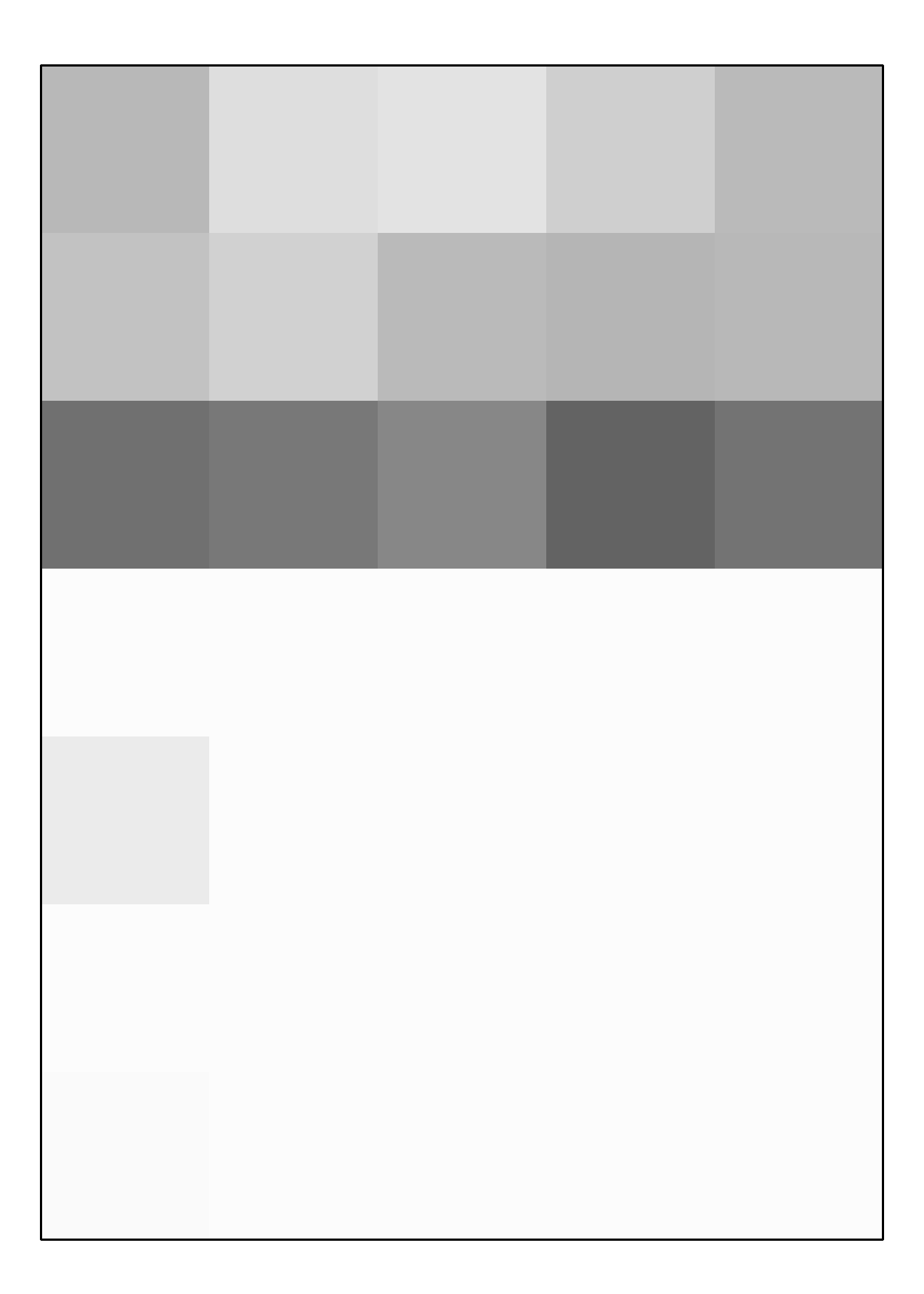}&\includegraphics[width=0.037\linewidth]{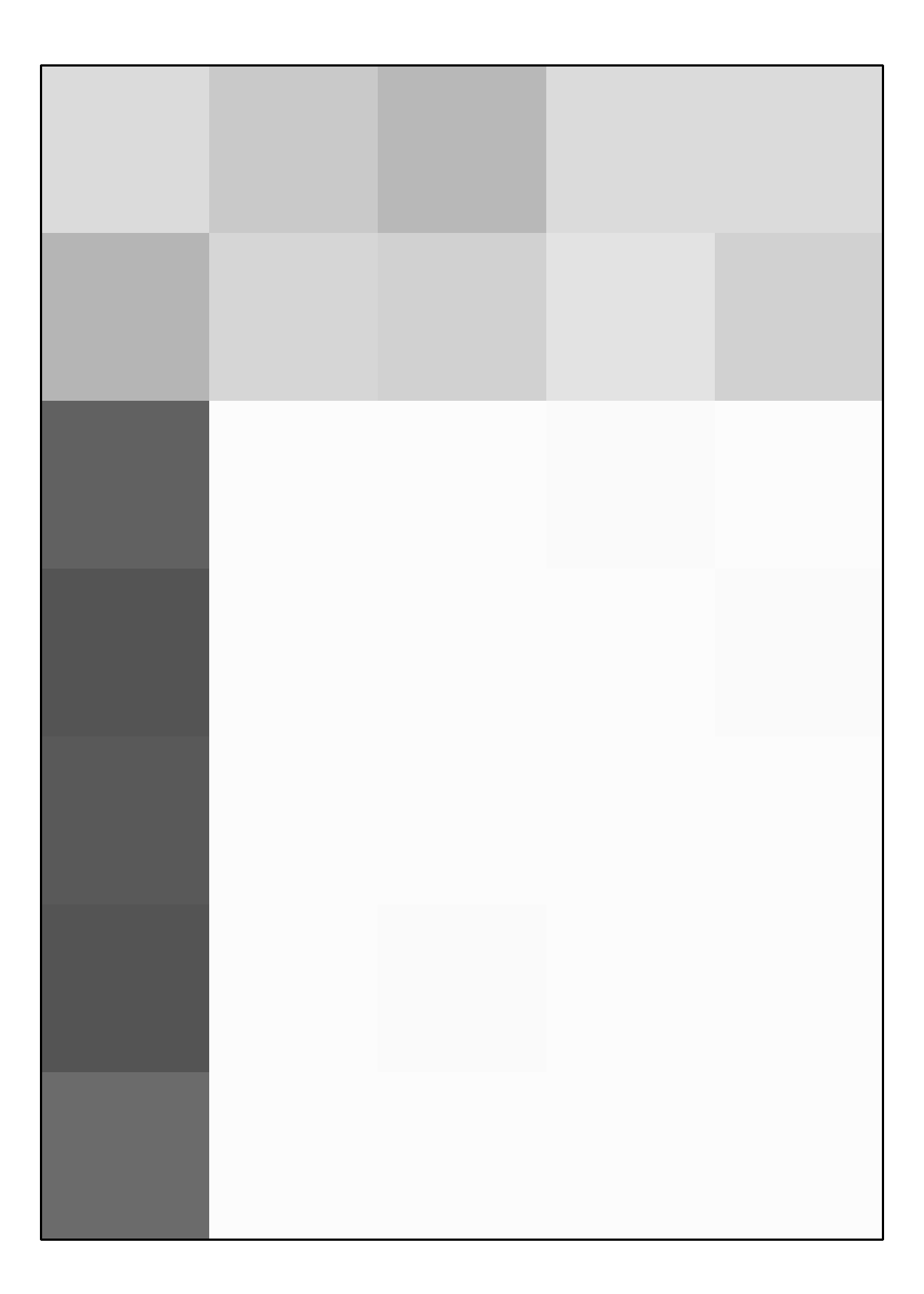}&\includegraphics[width=0.037\linewidth]{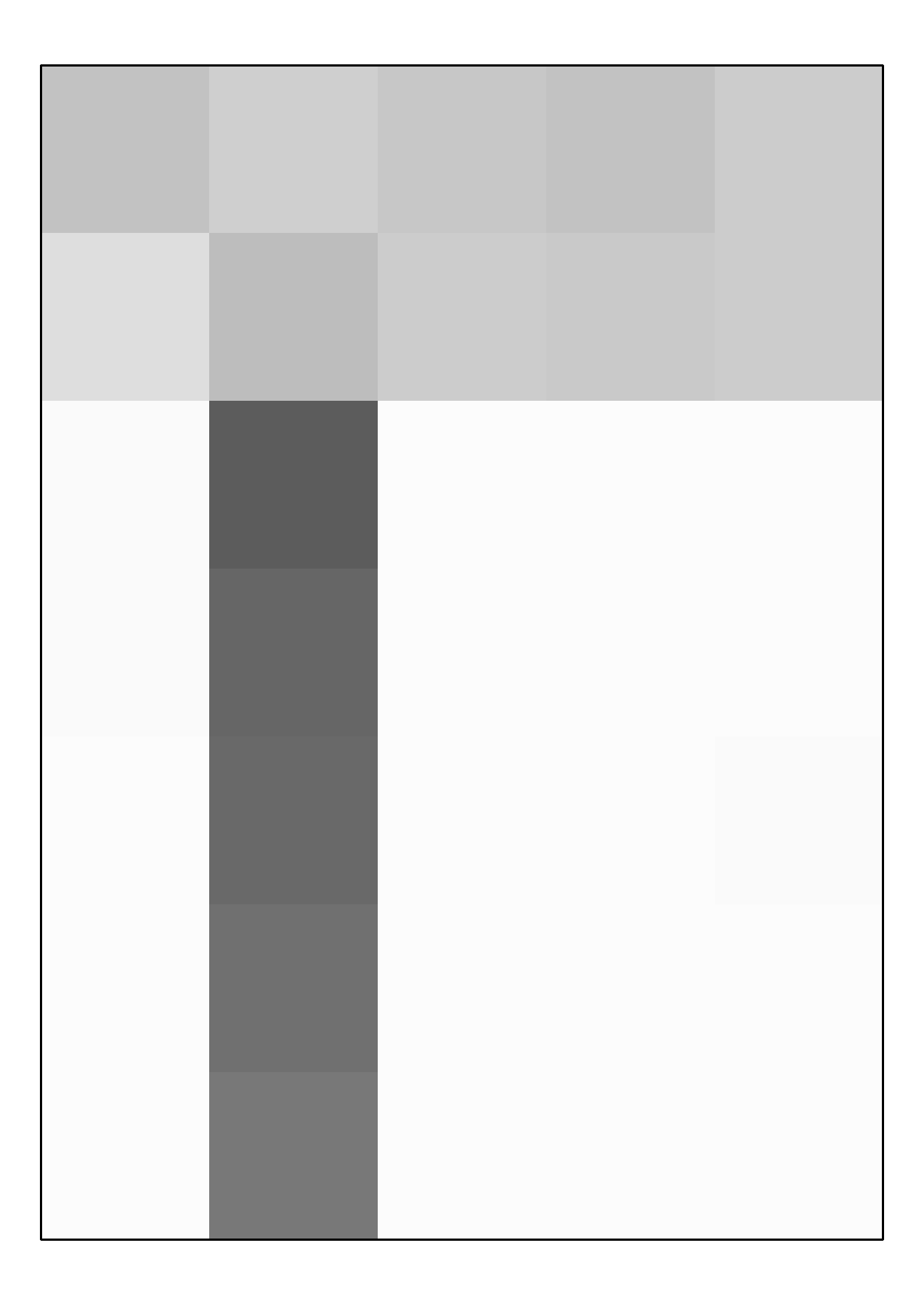}&\includegraphics[width=0.037\linewidth]{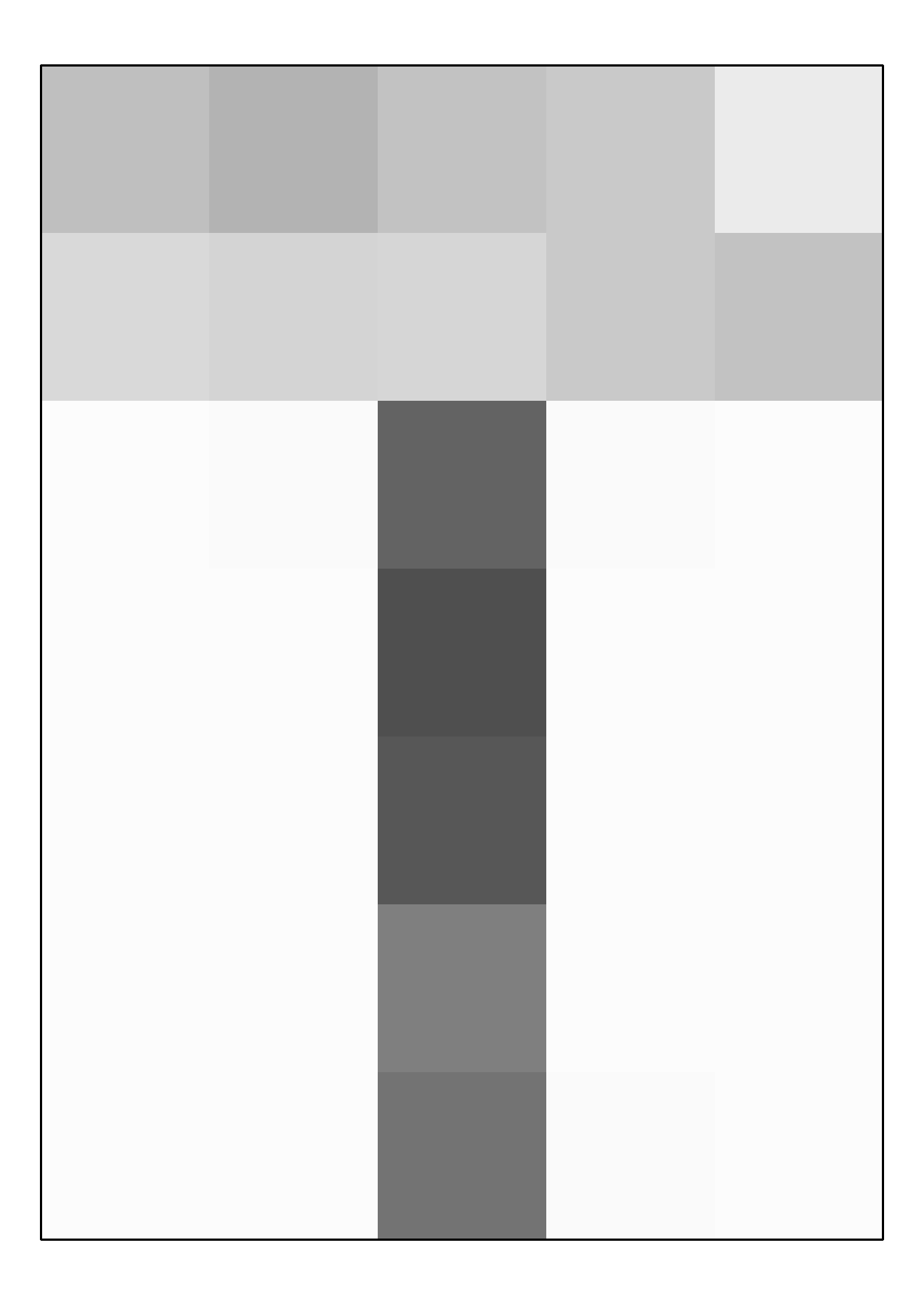}&\includegraphics[width=0.037\linewidth]{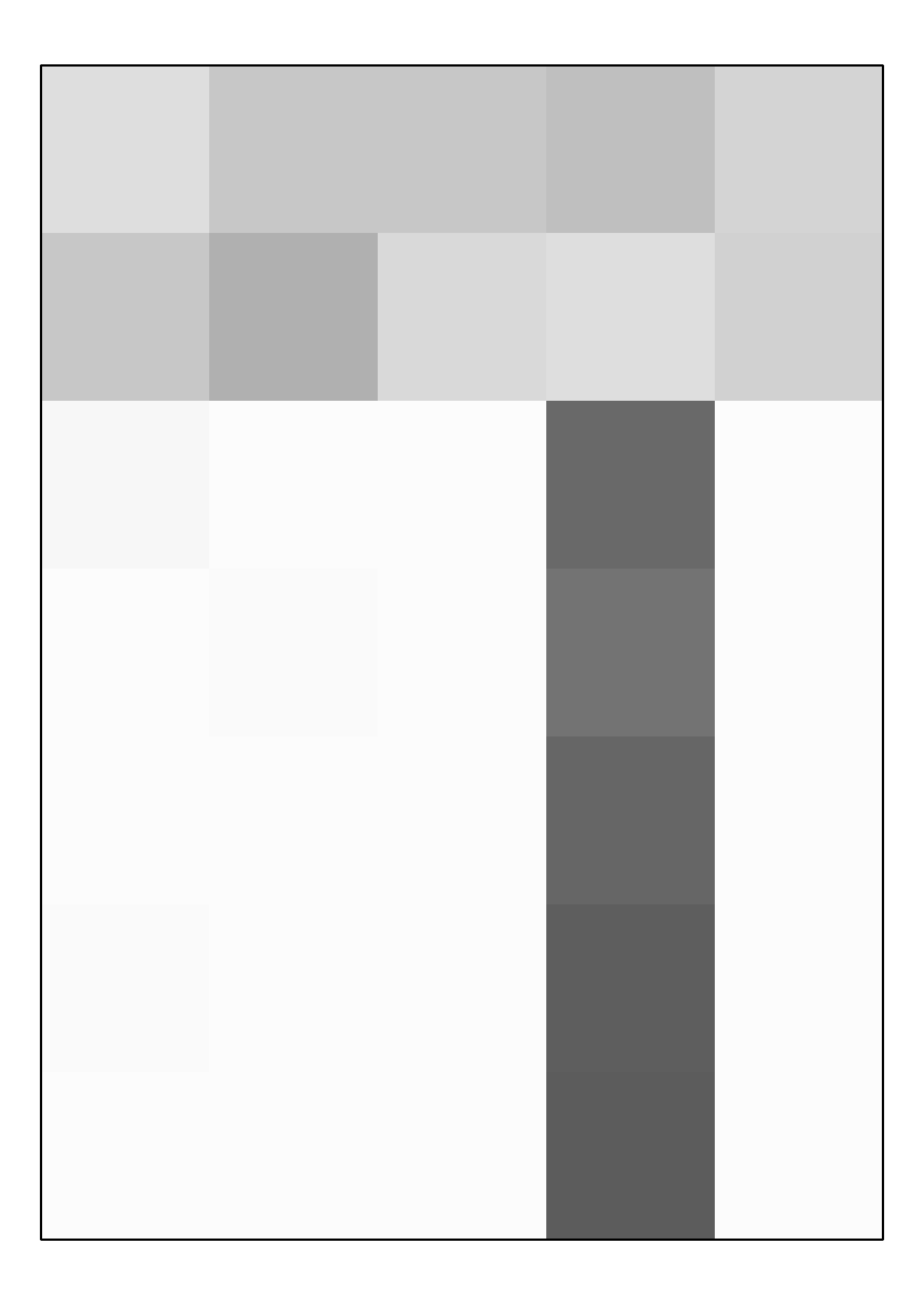}&\includegraphics[width=0.037\linewidth]{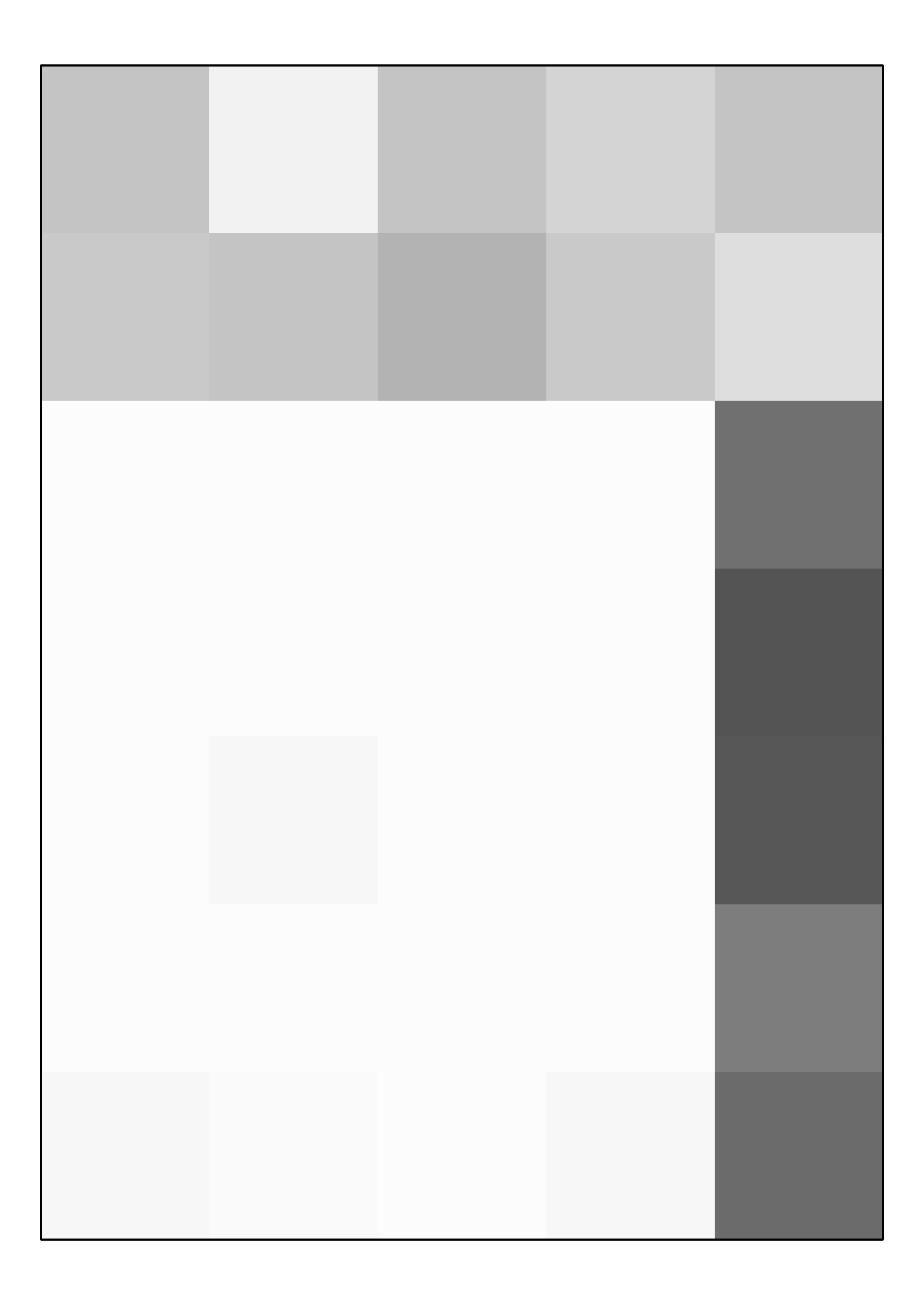}&\\
LDA11	&\includegraphics[width=0.037\linewidth]{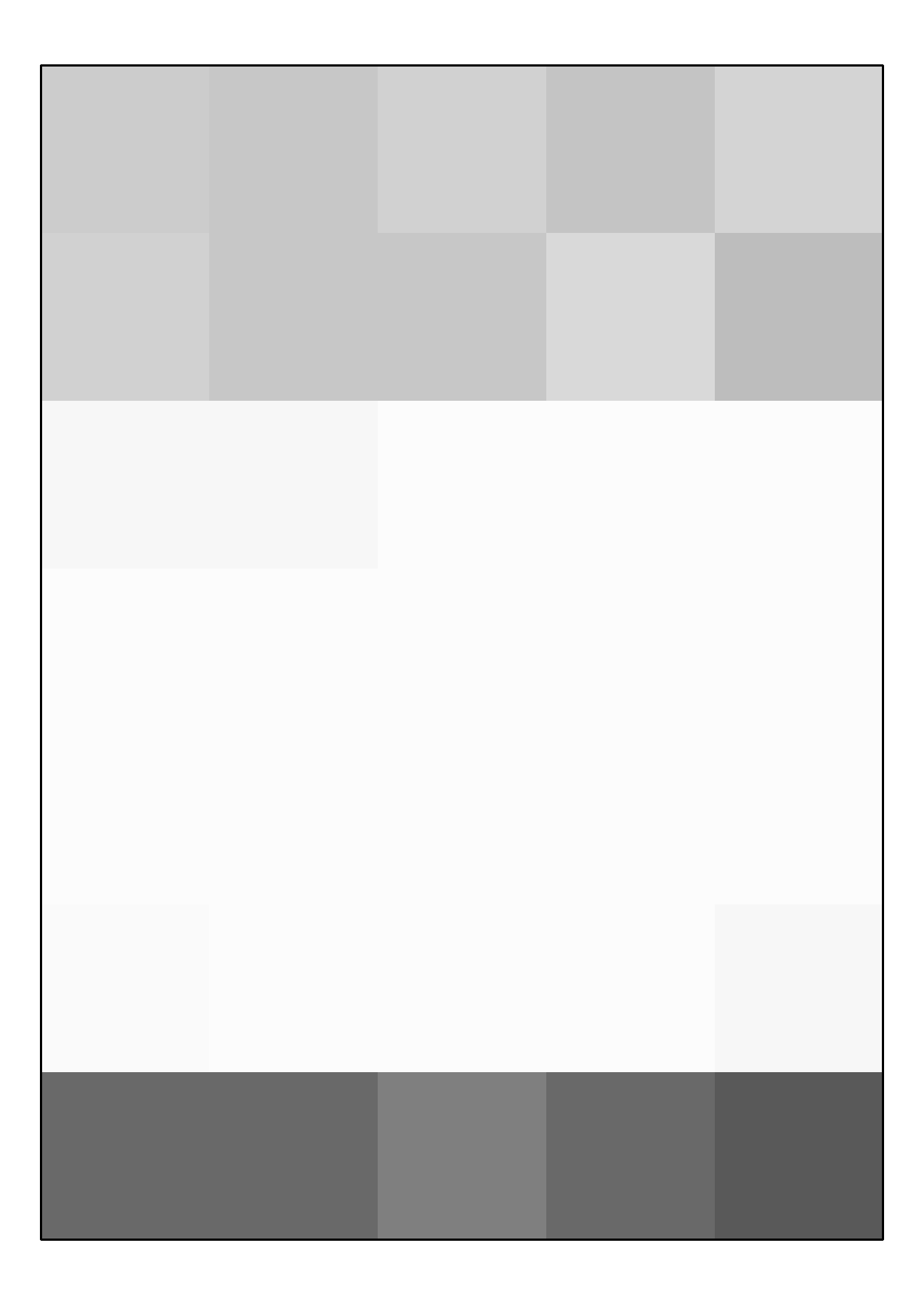}&\includegraphics[width=0.037\linewidth]{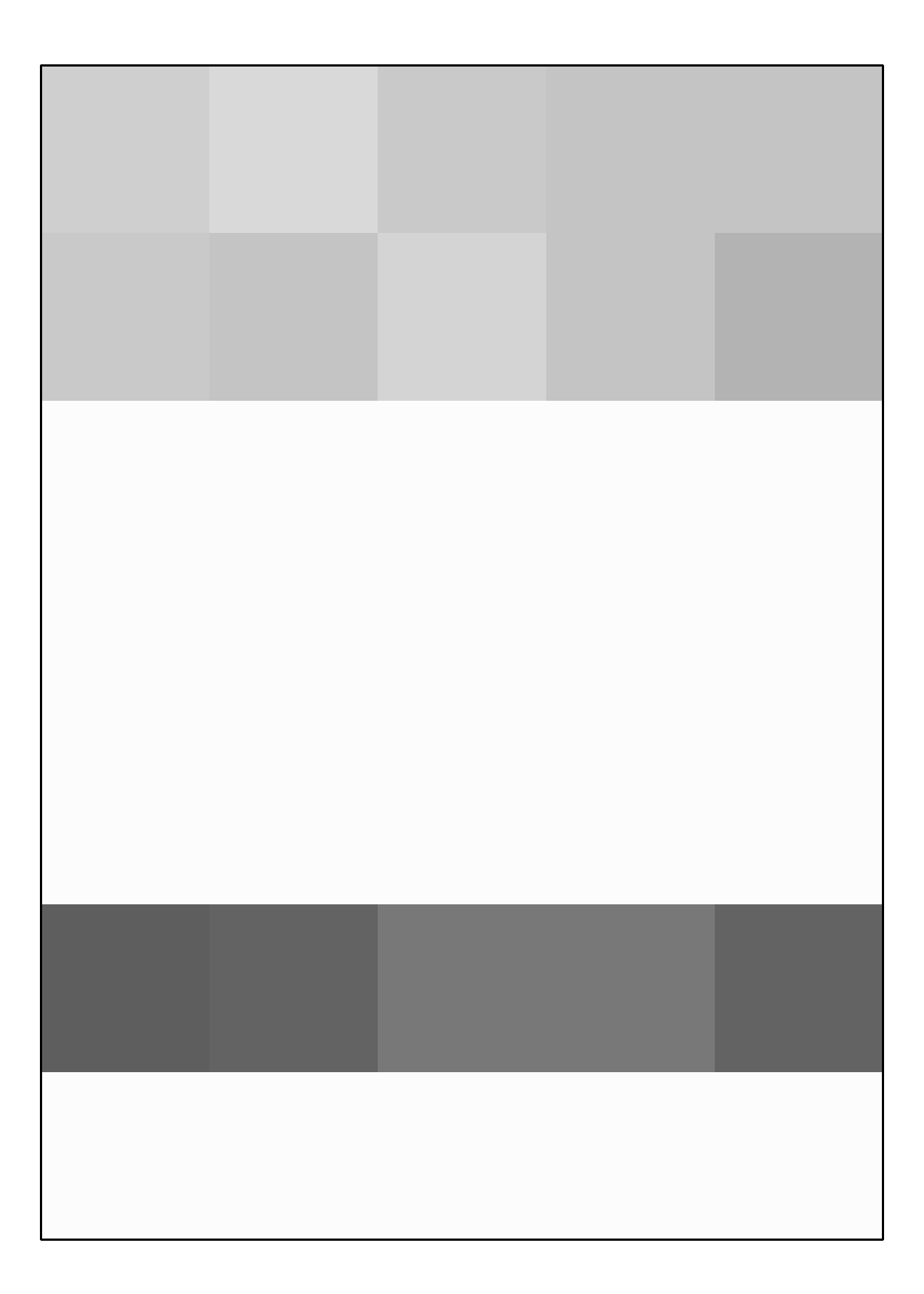}&\includegraphics[width=0.037\linewidth]{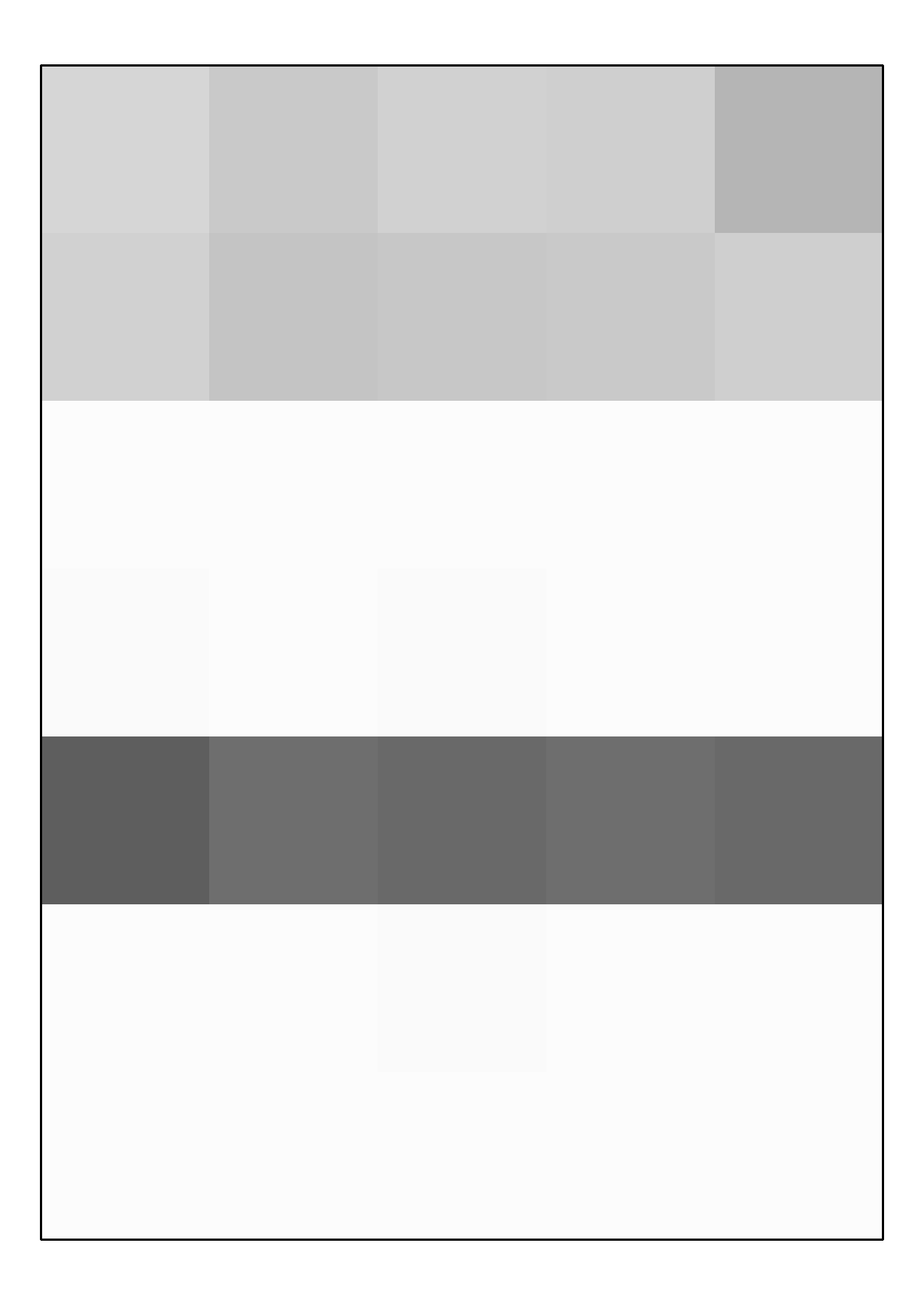}&\includegraphics[width=0.037\linewidth]{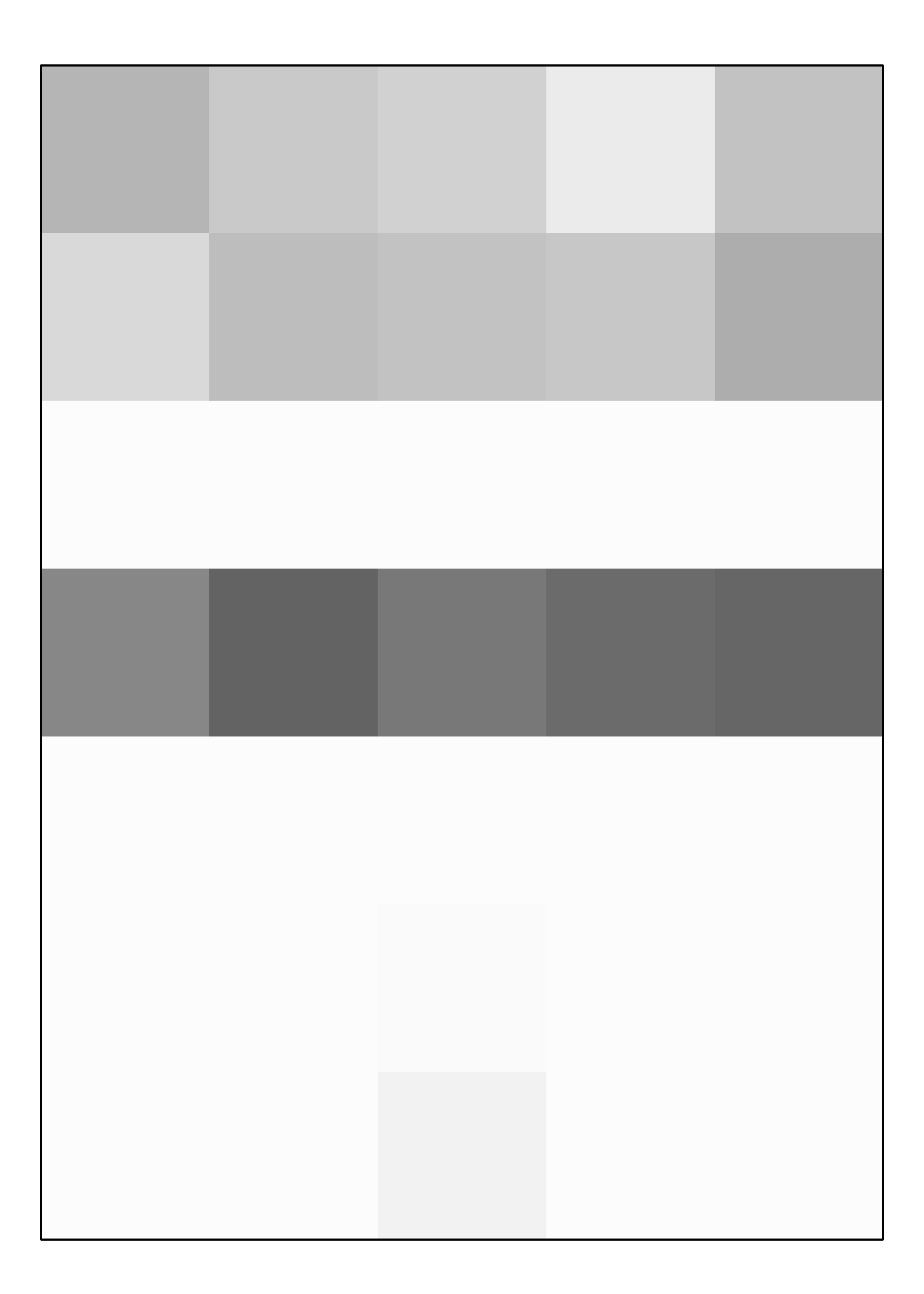}&\includegraphics[width=0.037\linewidth]{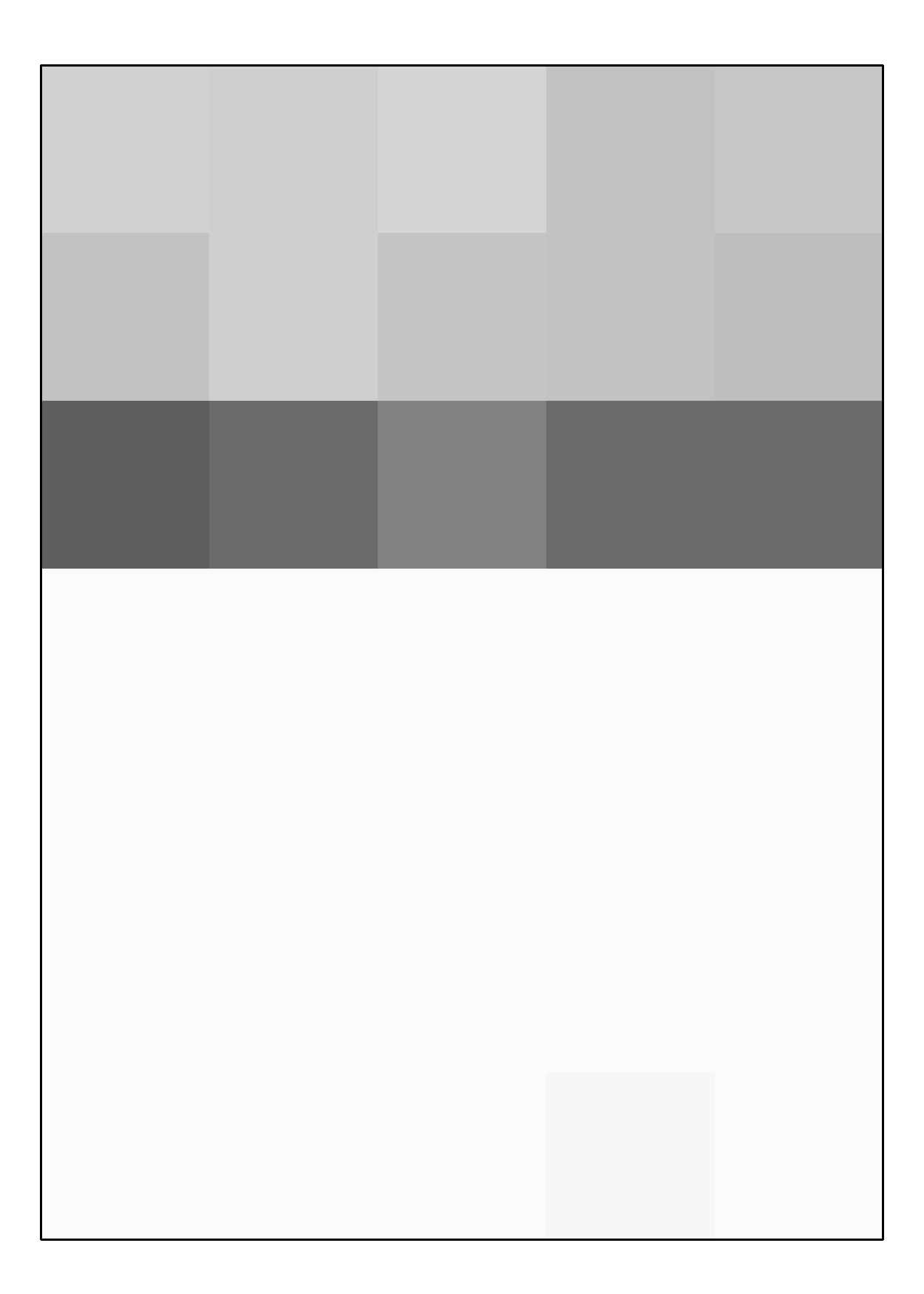}&\includegraphics[width=0.037\linewidth]{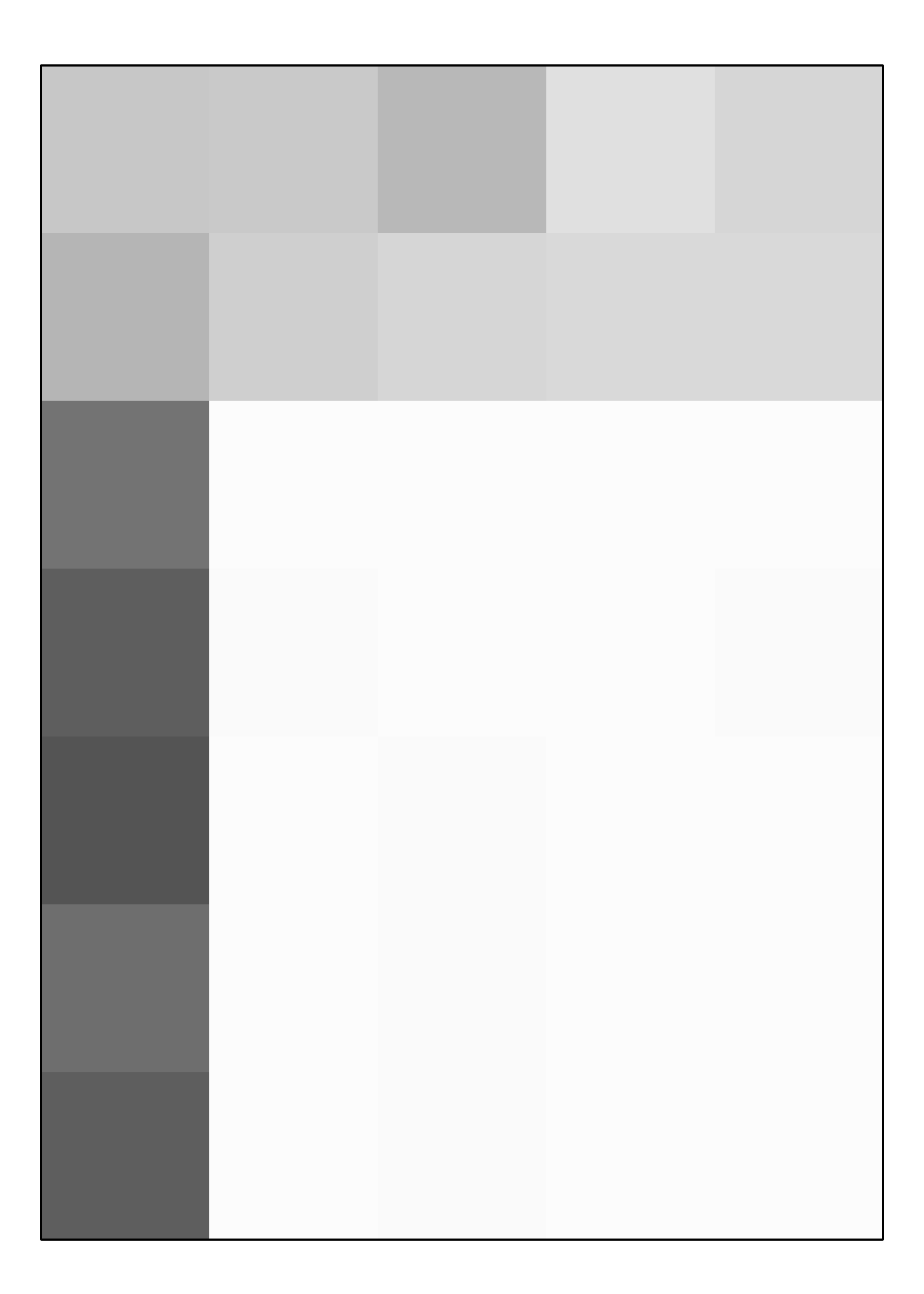}&\includegraphics[width=0.037\linewidth]{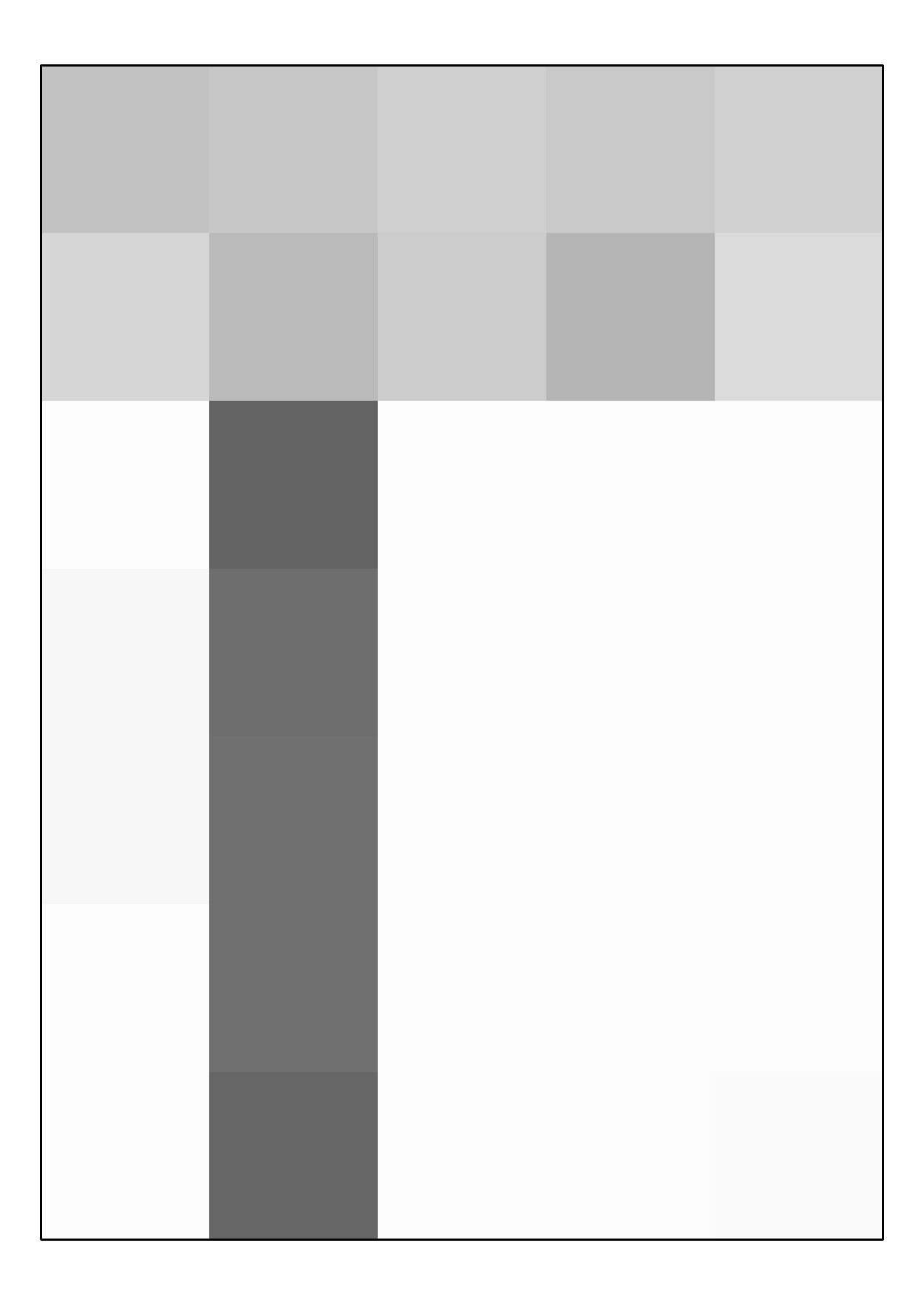}&\includegraphics[width=0.037\linewidth]{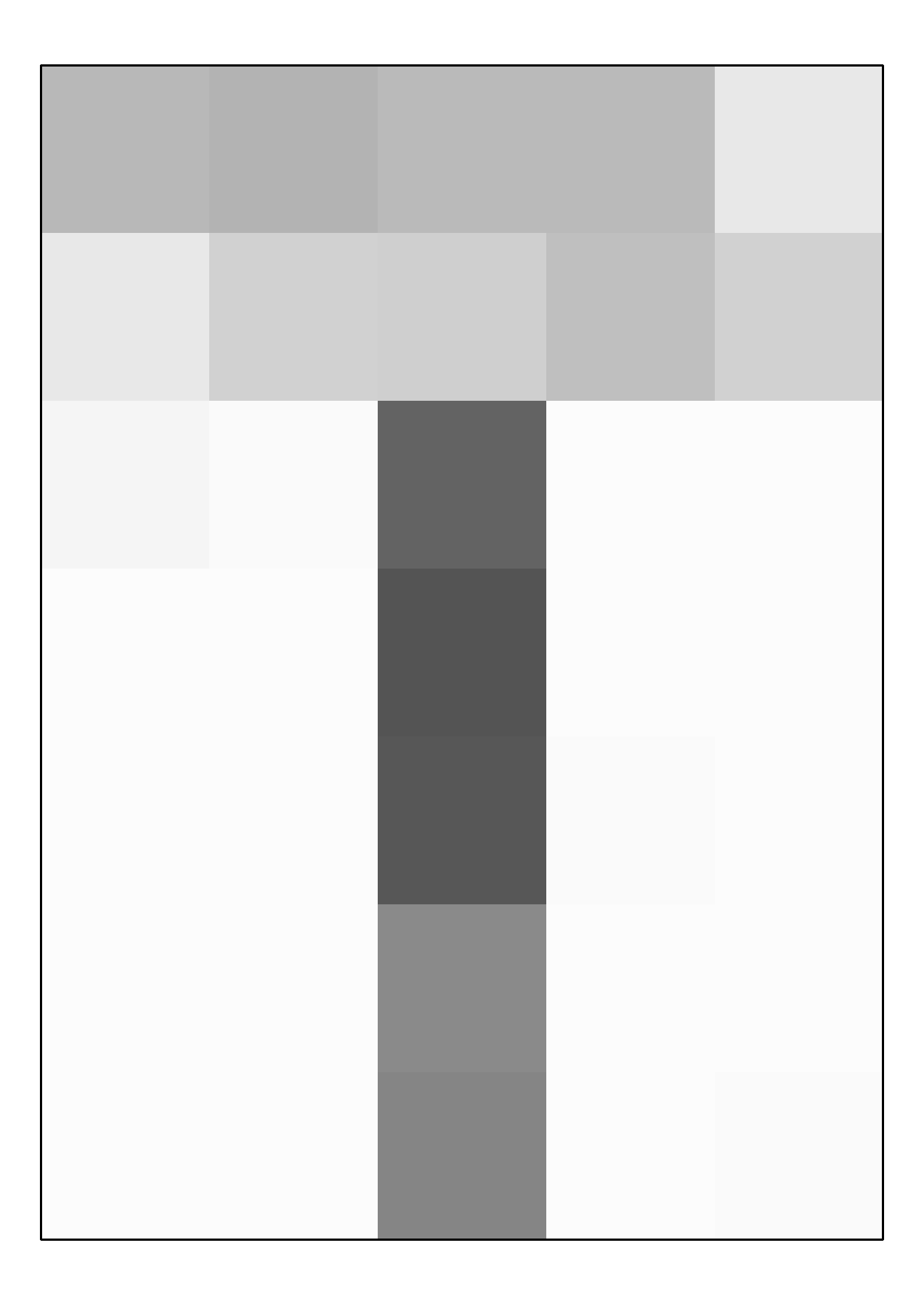}&\includegraphics[width=0.037\linewidth]{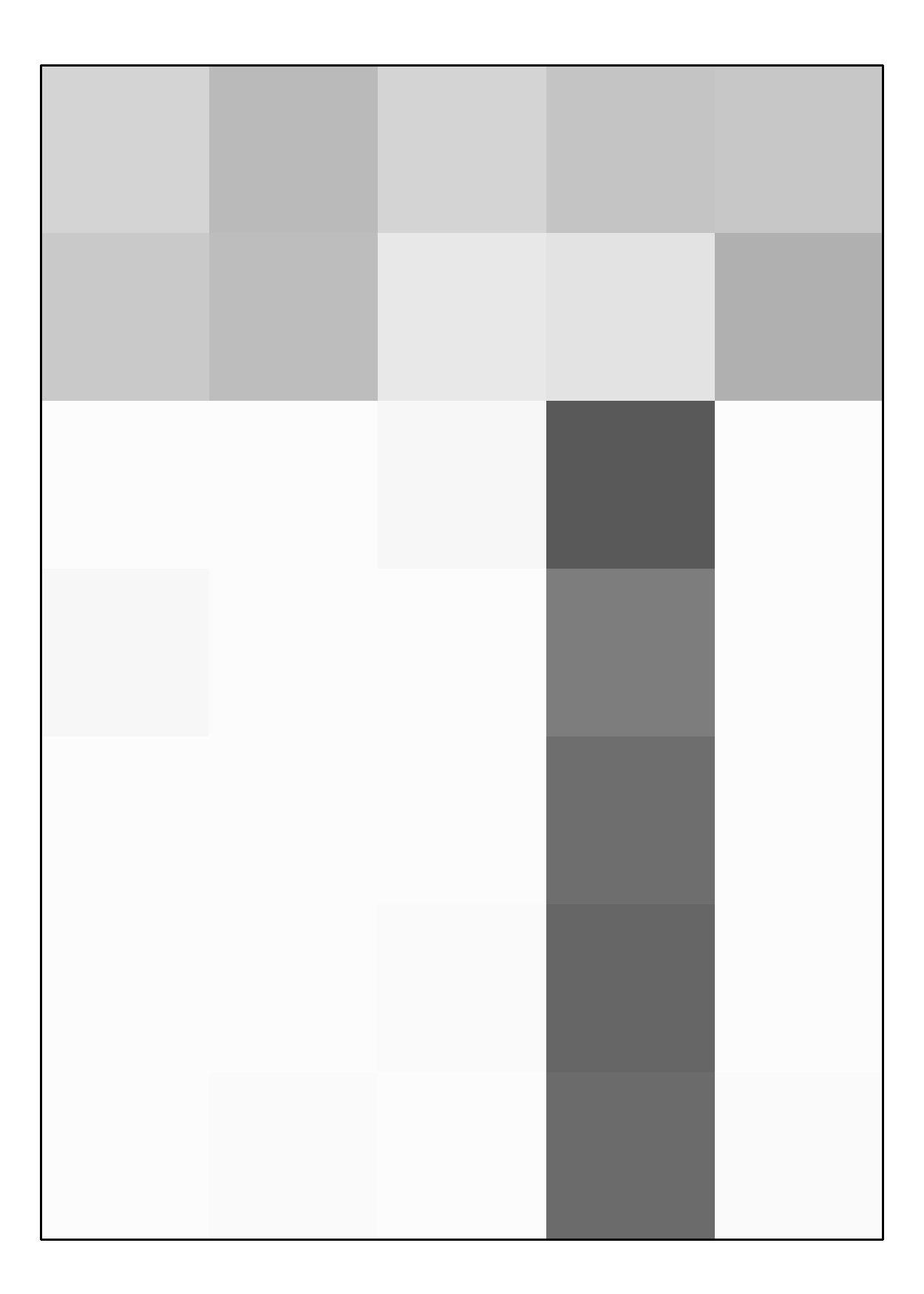}&\includegraphics[width=0.037\linewidth]{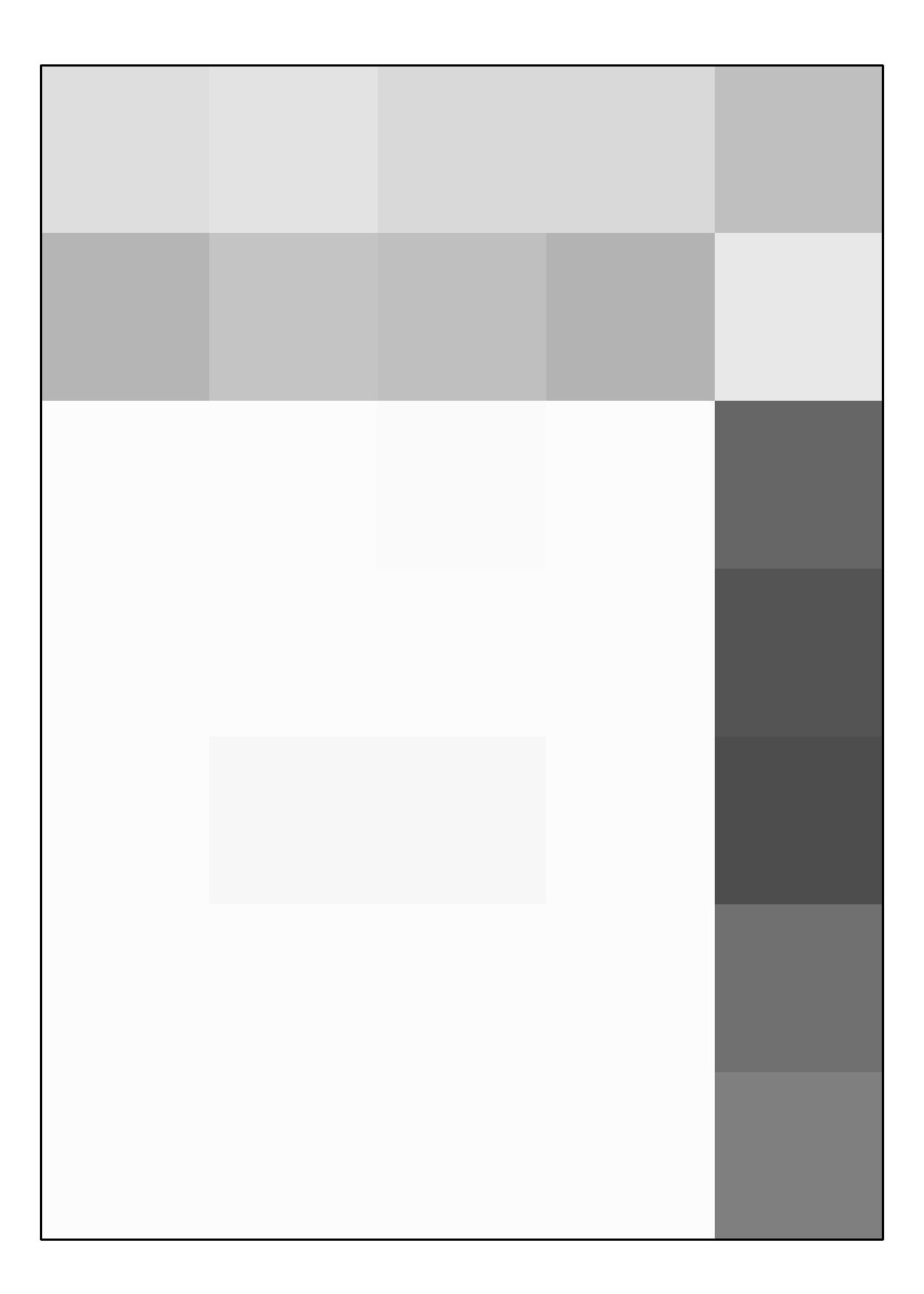}&\includegraphics[width=0.037\linewidth]{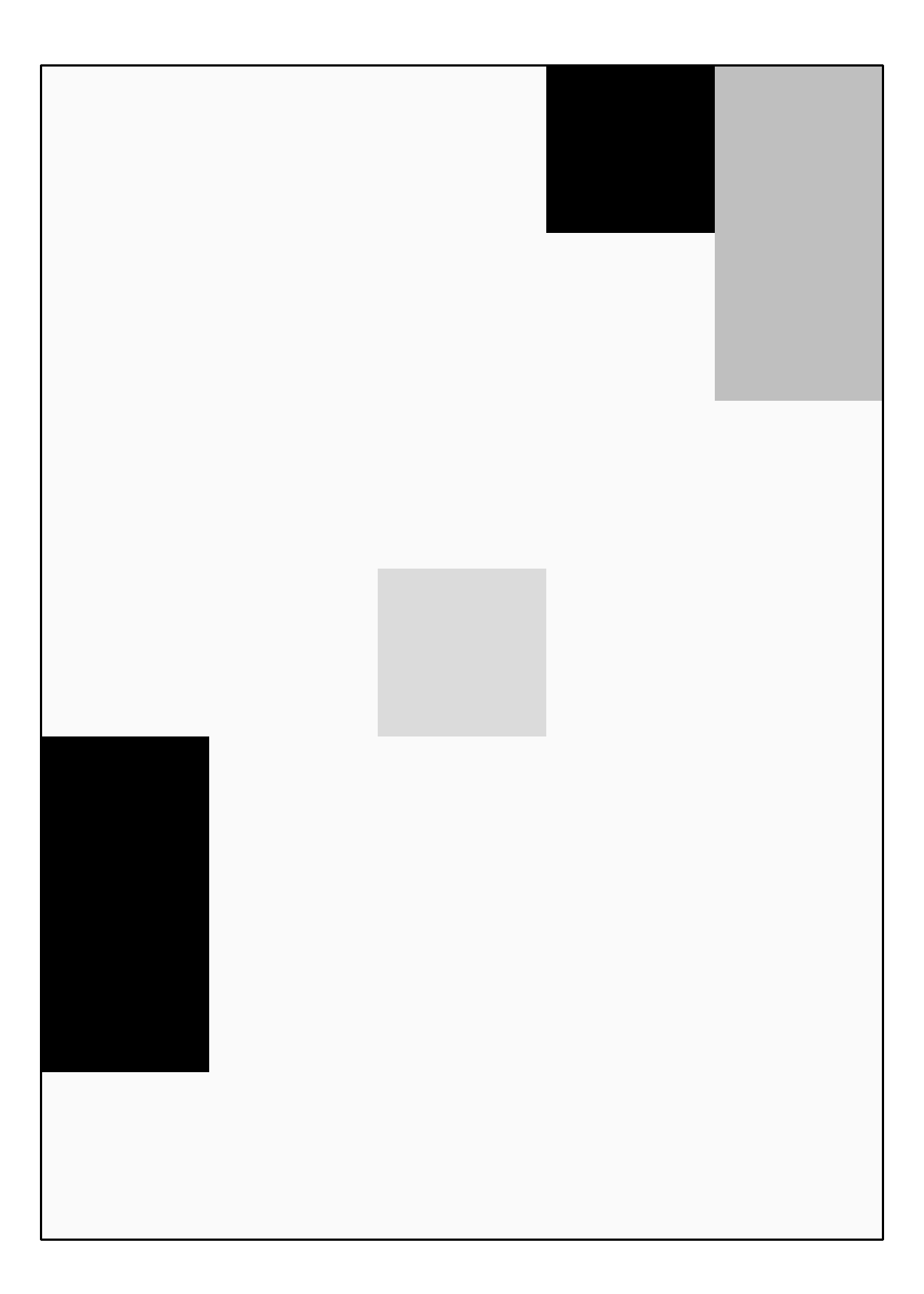}\\
    \end{tabular}
  \caption{ \label{fig:synthetic}Result of the simulation study. The first row shows the topics used to generate the data. The second, third, and fourth rows show the inferred topics from vsLDA, LDA10, and LDA11, respectively. For all models, we use an asymmetric prior $\boldsymbol{\alpha}$ over the per-document topic proportions.}
\end{figure*}

\section{Approximate Posterior Inference}
Deriving exact posterior distributions for the latent variables in vsLDA is intractable. We propose an MCMC algorithm to obtain posterior samples in order to make an approximate inference. Marginalizing over $\Phi, \psi$, and $\Theta$, the remaining latent variables in the joint likelihood are $\boldsymbol{z}, \boldsymbol{b}$, and $\boldsymbol{s}$ in equation \eqref{eqnm}. Given the word selector $\boldsymbol{s}$ and the observed data $W$, $\boldsymbol{b}$ is determined because $b_{di} = 1$ for all informative word tokens and $b_{di} = 0$ for all non-informative word tokens. Therefore, we sample only $\boldsymbol{z}$ and $\boldsymbol{s}$ through a collapsed Gibbs updating and a Metropolis updating relying on a Monte Carlo integration, respectively.

\textbf{Step 1 : Sampling $\boldsymbol{z}$} : Given $W$ and $\boldsymbol{s}$, we sample $z_{di}$ only for $d$ and $i$ such that $w_{di} = j$ and $s_j = 1$ (i.e., only for the word tokens taking the values in the informative word set). Letting $\boldsymbol{z}_{-di} = \{z_{d'i'}: d' \neq d$ or $i' \neq i\}$, the conditional distribution of $z_{di}$ given $\boldsymbol{z}_{-di}$, $\boldsymbol{s}$, and $W$ is

\begin{align}
\label{gibbs}
p(z_{di} = k | W, \boldsymbol{z}, \boldsymbol{s}) \propto (n_{d\cdot}^{k} + \alpha) \frac{n_{\cdot w_{di}}^{k} + \beta}{n_{\cdot \cdot}^{k} + \beta\sum_{j=1}^{V} s_j}
\end{align}

which depends only on the number of informative words and the topic assignments of the other informative word tokens. This is a generalization of updating step for topic assignment in typical LDA models where the vocabulary size is fixed as $V$ while it varies as $\sum_{j=1}^{V} s_j$ in our model.

\textbf{Step 2 : Sampling $\boldsymbol{s}$} : We let $\boldsymbol{z}^j = \{ z_{di}; w_{di} = j \}, \boldsymbol{z}^{-j} = \boldsymbol{z} \backslash \boldsymbol{z}^{j}$, and $\boldsymbol{s}^{-j} = \boldsymbol{s} \backslash {s}^{j}$. We update $s_j$ using a Metropolis step where $s_j^{proposed}$ is accepted over $s_j^{current}$ with a probability

\begin{align}
\mbox{Min} \left\{ 1, \frac{\int p(W, \boldsymbol{z}^{j}, \boldsymbol{z}^{-j}, s_j^{proposed}, \boldsymbol{s}^{-j}) p^*(\boldsymbol{z}^j)d\boldsymbol{z}^j }{\int p(W, \boldsymbol{z}^{j}, \boldsymbol{z}^{-j}, s_j^{current}, \boldsymbol{s}^{-j}) p^*(\boldsymbol{z}^j)d\boldsymbol{z}^j }  \right\}
\end{align}

where $p^*(\boldsymbol{z}^j) = p( \boldsymbol{z}^{-j}, s_j,  \boldsymbol{s}^{-j})$ is the conditional distribution of $\boldsymbol{z}^{j}$ given all the others. If proposed or current $s_j$ is 0, $ \boldsymbol{z}^j$ disappears in the joint likelihood as $p(W, \boldsymbol{z}^{-j}, s_j = 0, \boldsymbol{s}^{-j})$ and we do not need to marginalize over $\boldsymbol{z}^{j}$. If proposed or current $s_j$ is 1, marginalization over $\boldsymbol{z}^{j}$ does not yield a closed form and we rely on a Monte Carlo integration as follows.

\begin{align}
&\int p(W, \boldsymbol{z}^{j}, \boldsymbol{z}^{-j}, s_j = 1, \boldsymbol{s}^{-j})p^*(\boldsymbol{z}^{j}) d\boldsymbol{z}^{j} &\notag\\
&\quad{}= \frac{\sum_{u=1}^{U} p(W, \boldsymbol{z}^{j(u)}, \boldsymbol{z}^{-j}, s_j = 1, \boldsymbol{s}^{-j})}{U}&
\end{align}

where $\boldsymbol{z}^{j(u)}$ is $u$th sample obtained from $p^*(\boldsymbol{z}^{j})$ as in equation  \eqref{gibbs}. Once we obtain posterior samples of $\boldsymbol{s}$, inference about $\boldsymbol{s}$ is done through

\begin{align}\label{posS}
s_j = \mbox{argmax}_{B < t \leq T} p(s_j^{(t)} | W, Rest^{(t)} )
\end{align}

where $T$ is the total number of iterations and $B$ is the burn-in count.

\section{Simulation Study}

We first verify the correctness of our model with a synthetic dataset. We generate the synthetic corpus as follows. We start with thirty-five words in the vocabulary, design ten topics such that each topic has five topic (informative) words with 0.2 probability each and zero probability for all other words. Then we add a non-informative set with ten words, that do not appear in any of the topics, with 0.1 probability each. The first row in Figure \ref{fig:synthetic} shows these hand-crafted topics. As the figure shows, there are twenty-five informative words and ten non-informative words. Based on these topics, we generate 200 documents, 40 to 50 tokens each, with random topic proportions drawn from the Dirichlet distribution with a symmetric prior of 0.1. For each document, we set $\tau$ to 0.6, which means 60\% of word tokens are drawn from the topics, and 40\% word tokens are drawn from the non-informative set.

With this synthetic corpus, we trained vsLDA with ten topics and LDA with ten (LDA10) and eleven (LDA11) topics. For the hyperparameters, we place an asymmetric $\alpha$ prior over the document topic proportions, a symmetric $\beta$ prior over the topic-word distributions, and a symmetric $\gamma$ prior over the non-informative word distribution. These asymmetric $\alpha$ and symmetric $\beta$ priors can improve the performance of LDA compared to the widely used symmetric $\alpha$ and $\beta$ priors \cite{citeulike:6487302}. During the inference steps, we optimized these hyperparameters by using Minka's fixed point iteration \cite{minka} except $\gamma$ which we set to 1. Finally, we place Beta(1,1) priors over the hyperparameters $\lambda$ and $\tau$.

Figure \ref{fig:synthetic} shows the topics inferred by each model. The topics from LDA10 and LDA11 look less clear than the topics from vsLDA. By design, vsLDA infers the non-informative words and explicitly excludes them from the topics, so the resulting topics distribute all of the probability over the informative words, thereby discovering topics with clearer patterns. In this simulation, vsLDA exactly captures the top five words in each topic, and correctly identifies, using  equation \eqref{posS}, the set of non-informative words.
LDA10 finds the top five words in each topic quite well. However, every topic identified by LDA10 distributes some probability over the non-informative words, so the topics are not clearly defined by the five topic words. One interesting point of discussion is that typically LDA with asymmetric $\alpha$ priors are known to capture the common words of the corpus into a topic, so we expect that LDA11 would capture the non-informative words in its eleventh topic. However, topic number 11 in LDA11 actually captures an ambiguous distribution which has a sparse distribution over the words. However, if we adjust $\tau$ to be smaller than 0.3, LDA11 captures the non-informative words into one topic as well.

\section{Empirical Study}

\begin{table}[t]
  \centering
 
    \begin{tabular}{crrrc}
    Dataset & \# of docs &  \# of words & \# of tokens  & Stopwords \\ \hline
	20NG & 2,000 &  3,608 & 155,622& No \\
	NIPS& 1,740  & 2,613 & 104,069 & Yes \\
	SigGraph&  783 & 2,808 & 54,804 & No \\	
    \end{tabular}
  \caption{ \label{tbl:dataset}Dataset statistics. The {\it stopwords} column indicates whether stopwords were kept (yes) or removed (no).}
\end{table}

\begin{figure}[t]
	\centering
	
	\includegraphics[width=\linewidth]{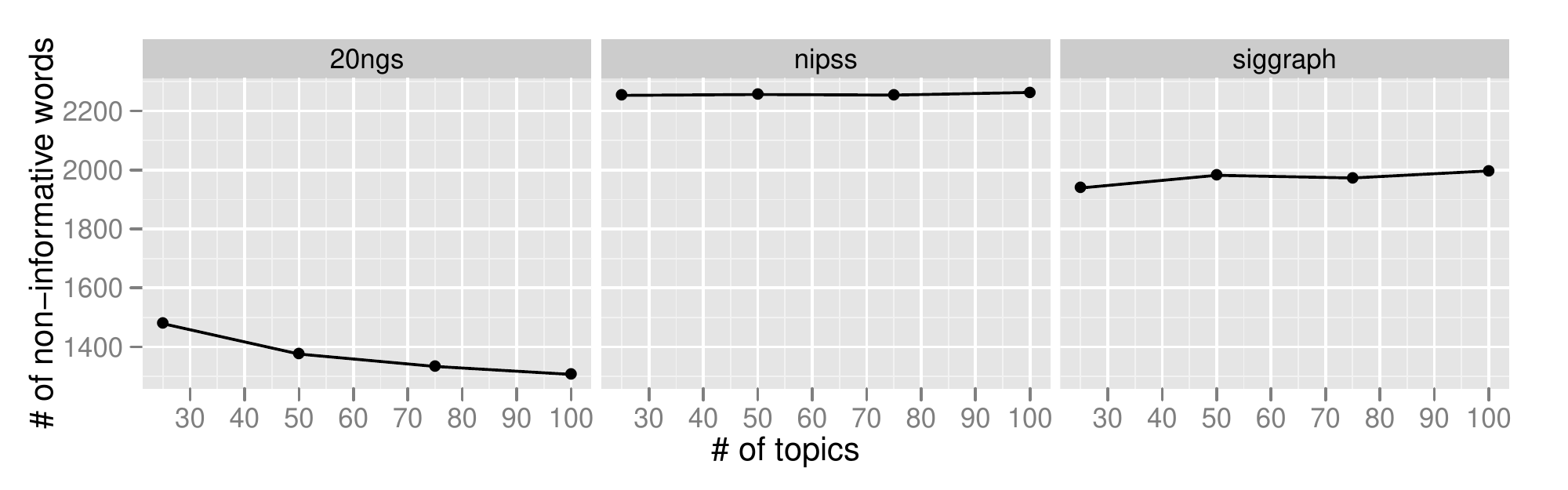}		

	\caption{\label{fig:numberofwords}	Change of the number of non-informative words over the number of topics ($K$).}
\end{figure}

In this section, we analyze three corpora for comparing vsLDA with two variants of LDA using various evaluation metrics. The first two are abstracts collected from the proceedings of the ACM SigGraph conferences (SigGraph) and the proceedings of the NIPS conferences (NIPS), and the third dataset is from the five comp subcategories from the 20 newsgroup corpus (20NG). To show the performance of vsLDA for diverse settings,  we test NIPS with stopwords kept and SigGraph and 20NG with stopwords removed. The detailed statistics of the three datasets are in Table \ref{tbl:dataset}. We compare three models: vsLDA, LDA with asymmetric priors for $\theta$ (asymLDA), and LDA with symmetric priors for $\theta$ (symLDA). Each model was run five times where each run includes 5,000 iterations with 3,000 burn-in samples and 100 iterations used as a thinning interval. Other parameters were optimized in the same way as the previous section.

\begin{figure}[t]
	\centering

	\subfigure[NIPS]{
	\includegraphics[width=0.3\linewidth]{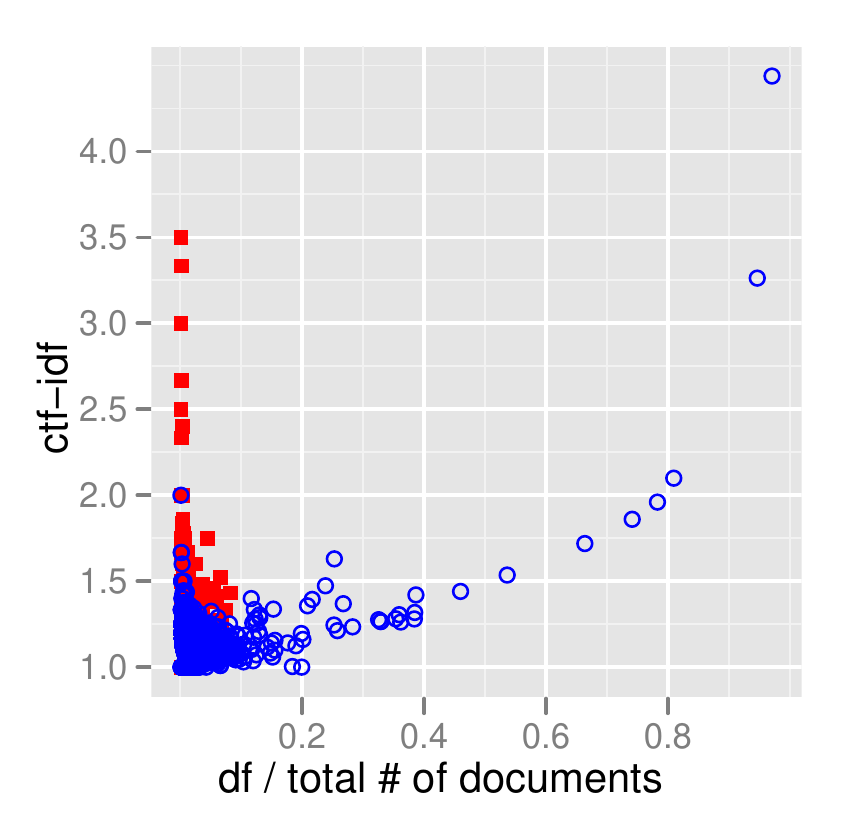}		
	}
	\subfigure[SigGraph]{
	\includegraphics[width=0.3\linewidth]{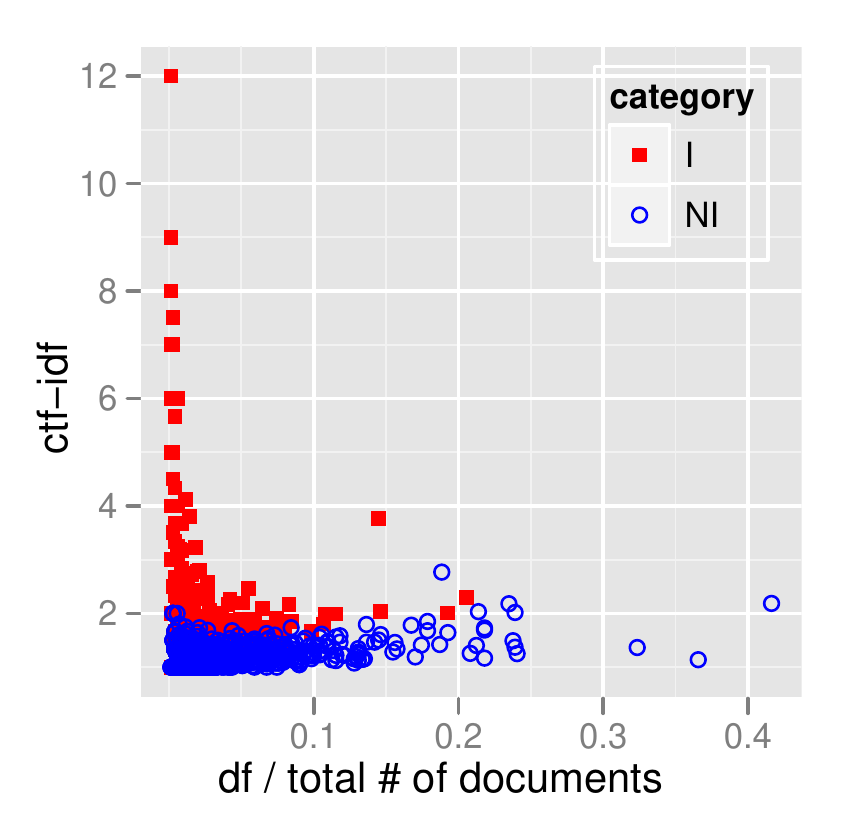}		
	}
	\subfigure[\label{fig:20ng}20NG]{
	\includegraphics[width=0.3\linewidth]{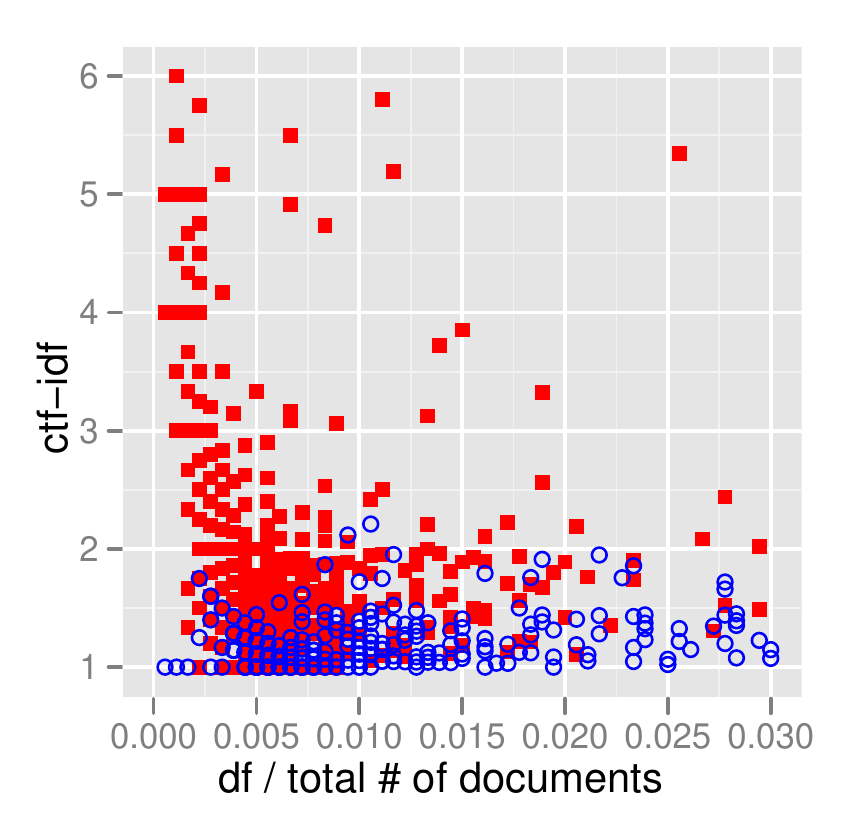}		
	}
	\caption{\label{fig:global}The scatter plot of {\it ctf-idf} versus relative df ({\it rdf = df / total \# of documents}) for the I words (red square) and the NI words (blue circle) inferred by vsLDA with $K = 50$ for each corpus. Generally, I words tend to have higher {\it ctf-idf} and lower {\it rdf} than the NI words.}
\end{figure}

\textbf{Characteristics of informative and non-informative words}
\quad{} We first describe the summary statistics of the informative (I) and the non-informative (NI) words found by vsLDA and explain the interesting patterns found. Figure \ref{fig:numberofwords} shows the pattern of how many words are found to be NI as we vary $K$, the number of topics. For NIPS and SigGraph, the number of NI words does not vary for $K$ of 25, 50, 75, and 100. For 20NG, the number noticeably decreases as $K$ increases. Further investigation is needed to explain this phenomenon, but as Table \ref{tab:holl} shows, log likelihood of heldout data follows a similar trend, so one conjecture is that the optimal number of topics is related to the number of NI words for a given corpus.
The proportion and the absolute number of the NI words clearly differ for each corpus. On average, vsLDA categorizes around 38\%, 70\%, and 86\% of the words as NI for 20NG, SigGraph, and NIPS, respectively.

To compare the characteristics of the NI and the I words, we compute three summary statistics of the words: (1) corpus term frequency ({\it freq}), (2) document frequency ({\it df}), and (3) corpus {\it tf-idf} ({\it ctf-idf}). Table \ref{tab:topwords} shows these statistics for ten words from 20NG and ten words from NIPS. To examine if any of the statistics are associated with word informativity, we ordered the words decreasingly by {\it freq} for 20NG and by {\it ctf-idf} for NIPS. Noting that there is no systematic pattern in the distribution of I and NI words in both orderings, we confirm that each statistic alone is not sufficient to distinguish the two classes of words inferred by vsLDA.

However, we found that the {\it ctf-idf} can be useful to quantify word informativity combined with the relative {\it df} ({\it rdf = df / total \# of documents}). Figure \ref{fig:global} shows a scatter plot of {\it ctf-idf} versus {\it rdf} for the I words (red square) and the NI words (blue circle) inferred from vsLDA. In particular, Figure \ref{fig:20ng} is a close-up of the lower-left corner where most of the words are located for 20NG. As shown in Figure 3, the I words tend to show higher {\it ctf-idf} and lower {\it rdf} than the NI words, suggesting that the I words are the ones that appear in a few documents (low {\it rdf}) but with high frequency (high {\it ctf-idf}. For SigGraph with $K = 50$, the average {\it ctf-idf} of I words is 2.26 and the average {\it ctf-idf} of NI words is 1.27. The other corpora at all levels of $K$ show the same pattern. As shown in Figure \ref{fig:20ng}, many of the words show low {\it rdf}, and classification of these words mainly depends on the high/low {\it ctf-idf}.

\begin{table*}[t]
	\centering
	
	\subtable[20NG]{
    \begin{tabular}{crrrcc}
   word &freq & df & ctf-idf & category \\ \hline
    subject & 1,855  & 1,715  & 1.08  & NI \\
    re    & 970   & 915   & 1.06  & NI \\
    windows & 918   & 356   & 2.58  & I \\
    writes & 822   & 653   & 1.26  & NI \\
    file  & 766   & 206   & 3.72  & I \\
    article & 686   & 537   & 1.28  & NI \\
    don't & 597   & 394   & 1.52  & NI \\
    scsi  & 592   & 89    & 6.65  & I \\
    program & 582   & 241   & 2.41  & NI \\
    drive & 569   & 199   & 2.86  & I \\
    \end{tabular}%
    }
    \subtable[NIPS]{
    \label{tab:nipstopwords}
\begin{tabular}{crrrc}
word  & freq  & df    & ctf-idf & category \\\hline
the   & 6,764  & 1,524  & 4.44  & NI \\
of    & 4,849  & 1,486  & 3.26  & NI \\
speech & 124   & 71    & 1.75  & I \\
localization & 19    & 11    & 1.73  & I \\
is    & 1,791  & 1,042  & 1.72  & NI \\
learning & 647   & 397   & 1.63  & NI \\
recurrent & 77    & 58    & 1.33  & I \\
hidden & 132   & 106   & 1.25  & I \\
feature & 66    & 53    & 1.25  & NI \\
can   & 493   & 396   & 1.24  & NI \\
\end{tabular}%
    }    
  \caption{\label{tab:topwords}Basic statistics for the I words and the NI words inferred by vsLDA with $K = 50$ for 20NG and NIPS. The words are ordered decreasingly by \textit{freq} (20NG) and by \textit{ctf-idf} (NIPS), and neither ordering shows a systematic pattern of word informativity.}
\end{table*}

In addition, Table \ref{tab:topwords} shows that the words normally categorized as stop words, such as ``the" and ``is" are correctly identified as NI, as are the words that do not distinguish topics, such as ``learning" and ``feature" in NIPS. 
We also found that the NIPS corpus contains 284 stopwords, and vsLDA identified 91\% of them on average as NI words.

\textbf{Held-out likelihood}
\quad{} vsLDA divides the vocabulary into I and NI, two mutually exclusive sets that are unpredictable given the basic word statistics. Now, we describe the performance of vsLDA using held-out likelihood which measures the model's predictive performance for an unseen document based on the trained parameters. We split the corpus into a training set containing 90\% of documents and a test set containing the rest. We compute held-out likelihoods using a left-to-right style sampler \cite{Wallach:2009p5474} with maximum a posteriori (MAP) estimators of parameters $\hat\phi$, $\hat\psi$, $\hat{\bf s}$, and $\hat\tau$. The average word likelihoods are shown in Table \ref{tab:holl}, and these results are consistent with the values reported in other studies of LDA with symmetric priors \cite{BoydGraber:2009p5267} and LDA with asymmetric priors \cite{citeulike:6487302}. Overall, the held-out likelihoods of vsLDA are higher than symLDA and comparable to asymLDA. It is worth noting again that vsLDA excludes the NI words, which make up 40\% to 80\% of the vocabulary, from the topics, and it still performs comparable to asymLDA which uses all of the words for the topics. These results suggest that including the NI words in forming the topics does not contribute to the predictive power of the model. To further test vsLDA, we manually set the stop words as the non-informative words, trained the model (stopword-vsLDA), and computed held-out likelihoods. These heldout likelihoods were lower than vsLDA and asymLDA. These results verify that variable selection must be done within the model in combination with the topics, rather than as a preprocessing step.

\begin{table}[t]
	\centering
	\subtable[20ng.comp]{
    		\begin{tabular}{cccc}
         	& vsLDA & asymLDA & symLDA \\\hline
 	   		25 & -7.04 &-7.00  &-7.35 \\
		    50    & -6.94  & -6.87  & -7.37  \\
		    75    & -6.88  & -6.82  & -7.35  \\
		    100   & -6.84  & -6.78  & -7.36  \\
    		\end{tabular}
	}	
	\subtable[SigGraph]{
    \begin{tabular}{cccc}
          & vsLDA & asymLDA & symLDA \\\hline
	    25    & -7.11  & -7.04  & -7.21  \\
	    50    & -7.09  & -7.02  & -7.25  \\
	    75    & -7.07  & -7.00  & -7.25  \\
	    100   & -7.06  & -6.99  & -7.27  \\
    \end{tabular}
    }
	\subtable[NIPS]{
    \begin{tabular}{ccccc}
          & vsLDA & asymLDA & symLDA & stopword-vsLDA* \\\hline
    25    & -6.28  & -6.25  & -6.34  & -6.32  \\
    50    & -6.28  & -6.25  & -6.43  & -6.34  \\
    75    & -6.28  & -6.25  & -6.40  & -6.34  \\
    100   & -6.28  & -6.25  & -6.43  & -6.35  \\
    \end{tabular}%
	
	}
    \caption{\label{tab:holl}$logP(W^{\text{test}}|W)/N^{\text{test}}$ for various values of $K$ for the three corpora. vsLDA performs comparable to asymLDA. For stopword-vsLDA, we manually set the NI words with stopwords. stopword-vsLDA performs comparable to symLDA but worse than vsLDA.}
    
\end{table}

\begin{figure*}[t!]
	\centering	
	\includegraphics[width=\linewidth]{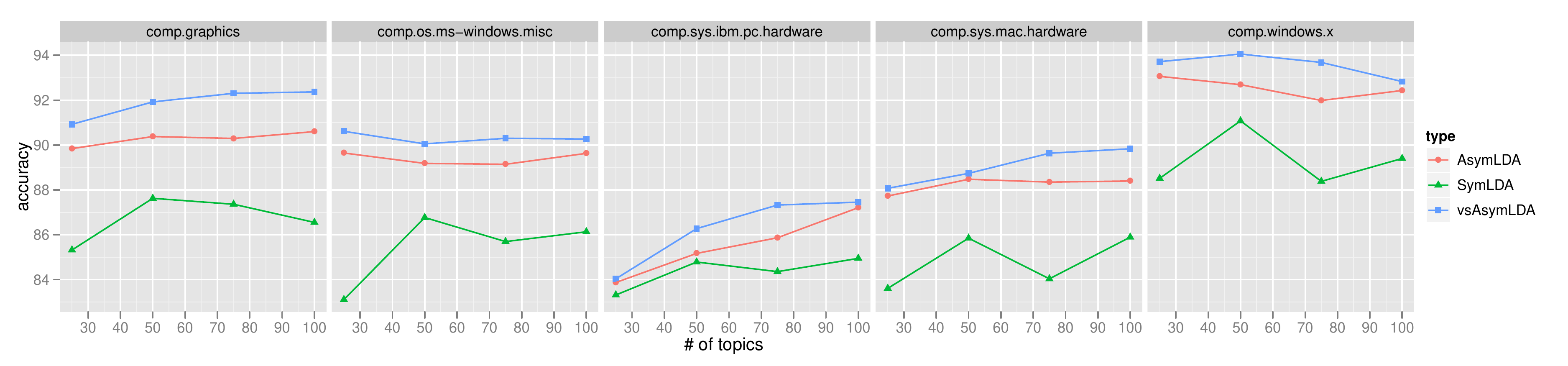}
	\caption{\label{fig:classification}Average classification accuracies using 20NG. vsLDA outperforms symLDA and asymLDA on document classification.}
\end{figure*}

\textbf{Classification}
\quad{}A topic model can be used for dimensionality reduction because it expresses each document as a finite mixture of topics. One way to verify the performance of a topic model is to perform classification tasks by using these reduced dimensions \cite{Blei:2003p4796,dag2006}. We use the five subcategories of the 20NG dataset and classify the documents into the subcategories. We use the libSVM toolkit with linear kernels, performing one-vs-all classification on each category with ten-fold cross validation.
Figure \ref{fig:classification} shows the average accuracies of the classification results. Overall, vsLDA performs better than all others with a small difference between vsLDA and asymLDA. We can conclude that vsLDA reduces each document into more discriminating subdimensions by excluding the non-informative words.

\begin{table}[t]

  \centering
  \subtable[K=50]{
    \begin{tabular}{lrrr}
          & 20ng.comp & NIPS  & SigGraph  \\\hline
	vsLDA & 3.12  & 2.49 & 3.22   \\
    asymLDA & 3.68  & 5.96  & 4.45     \\
    symLDA & 2.74 & 2.21  & 2.68 \\      
    \end{tabular}%
	}
  \subtable[K=100]{
    \begin{tabular}{lrrr}
     & 20ng.comp & NIPS  & SigGraph \\\hline
    vsLDA & 3.68  & 2.48  & 3.21  \\
   asymLDA& 4.40  & 7.38  & 5.77  \\
 symLDA & 3.00 & 2.20 & 2.69 \\
    \end{tabular}%
	}
  \caption{\label{tab:similarity}Average symmetric KL divergence between best matching topic pairs. vsLDA shows similar average divergences compared to symmetric LDA despite its asymmetric priors.}
\end{table}

\textbf{Similarity between multiple MCMC outputs with best matching algorithm}
\quad{} LDA with asymmetric priors tend to generate highly skewed distributions \cite{citeulike:6487302} where the model will capture several major topics well, but the other topics may be highly inconsistent over multiple MCMC outputs. In the experiments presented here, for instance, five major topics occupy more than 50\% of word tokens in the corpus. This may pose a problem for cases where the inferred topics $\hat\phi$, not just the topic assignments for the word tokens, are important.
Variation of information (VI) is one metric to evaluate the performance of clustering \cite{citeulike:3906686,citeulike:6487302}, but VI is based on mutual information of the topic assignments of tokens, so the major topics of the asymmetric models will overtake the VI metric, thereby masking the inconsistencies of the minor topics.

In order to better measure the consistency of the model with respect to the topics $\hat\phi$, we propose a new similarity metric based on the best matching algorithm. First, based on the inferenced $K$ maximum a posterior (MAP) $\hat{\phi}$s for each MCMC output, we find the best matching $K$ pairs that minimize the sum of symmetric KL-divergence with the Hungarian algorithm \cite{Goldberger:2003:EIS:946247.946585,springerlink:10.1007}. If the model generates consistent topics over multiple runs, then the sum of the divergences will also be minimized. Table \ref{tab:similarity} shows the average divergences between the best matching pairs, and it shows that vsLDA finds more consistent topics than asymLDA and comparable results with symLDA. The inconsistencies of asymLDA can be attributed to the minor topics for which the corpus does not exhibit regular word-topic patterns. Although vsLDA may also generate skewed distributions, vsLDA would use the NI category to exclude the words that do not exhibit clear topic patterns, so the resulting topics are more consistent and robust to the initializations of multiple MCMC runs.

We also measure the consistency of vsLDA by looking at the NI word sets over multiple runs and computing the Jaccard's coefficient, which measures the degree of overlap of two sets by dividing the intersection by the union. Although we do not know the `ground truth' of the NI word set, we certainly do not expect it to change for each run. The Jaccard's coefficients for multiple MCMC runs are, on average, 0.83, 0.96, and 0.93 for 20NG, NIPS, SigGraph, respectively, and these values represent high consistencies over multiple runs.

\section{Discussion}
We developed a variable selection model for LDA which selects a subset of the vocabulary to better model the topics. We were motivated by the curiosity about the usual practice of using the entire vocabulary to model the topics and the ad-hoc nature of the preprocessing steps to reduce the vocabulary size. Our model, vsLDA, explicitly selects the non-informative words to exclude from the vocabulary, simultaneously with the inference of the latent topics. By only using the words that help, not hinder, the process of inferring the topics, our model combines the advantages of LDA with symmetric priors and LDA with asymmetric priors. One future direction for vsLDA is to apply it to online learning \cite{Hoffman_Blei_Bach_2010,Yao:2009}. Typically, in an online learning situation, the vocabulary size gets larger as more data become available, but we cannot use the entire vocabulary because it monotonically increases \cite{Manning:2008:IIR:1394399}. By using vsLDA we can control the effective size of the vocabulary. Also, vsLDA can be used for object recognition, image segmentation \cite{splda2007,Zhao:2010:IST:1888150.1888211}, or collaborative filtering \cite{hofmann2004,citeulike:3998703} because vsLDA finds topics with more discriminative power. With vsLDA, we showed one way of incorporating variable selection into LDA and improving the results, so the natural next step would be to incorporate variable selection into other topic models \cite{citeulike:432492,dag2006,authortopic2004,guo2010topic} for improved results.

\bibliographystyle{plain}
\bibliography{vslda}

\end{document}